\documentclass{article}

\usepackage{arxiv}

\usepackage[utf8]{inputenc} % allow utf-8 input
\usepackage[T1]{fontenc}    % use 8-bit T1 fonts
\usepackage{hyperref}       % hyperlinks
\usepackage{url}            % simple URL typesetting
\usepackage{booktabs}       % professional-quality tables
\usepackage{amsfonts}       % blackboard math symbols
\usepackage{nicefrac}       % compact symbols for 1/2, etc.
\usepackage{microtype}      % microtypography
\usepackage{lipsum}		% Can be removed after putting your text content
\usepackage{graphicx}
\usepackage{natbib}
\usepackage{doi}

\usepackage{url,hyperref,lineno,microtype, subcaption}
\usepackage[onehalfspacing]{setspace}

\usepackage{textcmds}
\usepackage{color, colortbl}
\definecolor{Gray}{gray}{0.9}
\usepackage{multirow}
\usepackage{multicol}
\usepackage{tabularx}
\usepackage{siunitx}
\usepackage{graphicx}
\usepackage{xcolor}
\usepackage{caption}

\usepackage[export]{adjustbox}
\usepackage{comment}

\usepackage{xspace}
\newcommand*{\cf}{cf.\@\xspace}

\usepackage[noabbrev, capitalize, nameinlink]{cleveref}

\title{Improving HRI through robot architecture transparency}

\author{Lukas Hindemith\\
        CITEC\\
        Bielefeld University\\
        33619 Bielefeld\\
        Germany \\
        \texttt{lhindemith@techkfak.uni-bielefeld.de}
        \And
        Anna-Lisa Vollmer\\
        CITEC\\
        Bielefeld University\\
        33619 Bielefeld\\
        Germany \\
        \texttt{avollmer@techkfak.uni-bielefeld.de}
        \And
        Christiane B. Wiebel-Herboth\\
        Honda Research Institute Europe\\
        63073 Offenbach\\
        Germany\\
        \texttt{christiane.wiebel@honda-ri.de}
        \And
        Britta Wrede\\
        CITEC\\
        Bielefeld University\\
        33619 Bielefeld\\
        Germany \\
        \texttt{bwrede@techkfak.uni-bielefeld.de}
}

\hypersetup{
pdftitle={Improving HRI through robot architecture transparency},
pdfsubject={Human-Robot Interaction},
pdfauthor={Lukas Hindemith},
pdfkeywords={explainable robots, human-robot interaction, mental models, visual programming, control architecture, user study},
}

\begin{document}
\maketitle

\begin{abstract}
	In recent years, an increased effort has been invested to improve the capabilities of robots.
    Nevertheless, human-robot interaction remains a complex field of application where errors occur frequently. The reasons for these errors can primarily be divided into two classes. Foremost, the recent increase in capabilities also widened possible sources of errors on the robot's side. This entails problems in the perception of the world, but also faulty behavior, based on errors in the system. Apart from that, non-expert users frequently have incorrect assumptions about the functionality and limitations of a robotic system. This leads to incompatibilities between the user's behavior and the functioning of the robot's system, causing problems on the robot's side and in the human-robot interaction. While engineers constantly improve the reliability of robots, the user's understanding about robots and their limitations have to be addressed as well. In this work, we investigate ways to improve the understanding about robots. For this, we employ \emph{FAMILIAR -- FunctionAl user Mental model by Increased LegIbility ARchitecture}, a transparent robot architecture with regard to the robot behavior and decision-making process. We conducted an online simulation user study to evaluate two complementary approaches to convey and increase the knowledge about this architecture to non-expert users: a dynamic visualization of the system's processes as well as a visual programming interface. The results of this study reveal that visual programming improves knowledge about the architecture. Furthermore, we show that with increased knowledge about the control architecture of the robot, users were significantly better in reaching the interaction goal. Furthermore, we showed that anthropomorphism may reduce interaction success.
\end{abstract}

% keywords can be removed

\section{Introduction}
Advancements in robot capabilities have facilitated the transition of application fields from mainly static environments to more dynamic ones. Not only did the environment change, but also the involvement of humans~\citep{sendhoff2020cooperative}. Thereby, scenarios emerged where the robot cannot pre-plan all actions in advance to accomplish a certain goal. Instead, only close interactions with a human partner allow the robot to determine the subsequent action. Moreover, not only experts are supposed to interact with a robot, but also \emph{na\"{i}ve users}. The term \emph{na\"{i}ve user} in the following denotes a human user without computer science background who is unfamiliar in interaction with a robotic system. However, we assume they live in a society where almost everyone is exposed to technology to some degree, such as computers or smartphones.

Although robots are capable of interacting with na\"{i}ve users, problems occur regularly~\citep{johnson2009autonomy, tsarouhas2016mission}. These can be attributed to both, robots and users. More functionality on the robot side increases the potential for errors. To perceive the world, various hardware sensors need to work reliably, several modules need to communicate with each other and data needs to be transferred \citep{brooks2017human}. The behavior of a robot also depends on successful perceptions of the environment, working actuators, and a correct design by engineers \citep{steinbauer2012survey}. Besides errors in the robot system, users have a tremendous influence on the performance of a robot. Especially in close human-robot interactions and cooperation, the successful behavior of a robot highly depends on the correct input by the user. A more detailed failure taxonomy can be found in \citep{honig2018understanding}.

Na\"{i}ve users are often unaware of the functionality and limitations of a robot. This can be ascribed to the \emph{mental model} a na\"{i}ve user has about the robot. According to~\citep{staggers1993mental}, a mental model is defined as a cognitive framework of internal representations humans build about things they interact with. When a person interacts with an artifact for the first time, an initial representation is built and continuously updated while interacting with it. This is influenced by prior experiences with other artifacts and expectations towards them. In the case of robots, people relate to experiences with other people \citep{nass1994computers}. Additionally, expectations about robots, formed by movies and media, may lead a na\"{i}ve user's mental model further away from reality. Consequently, a non-functional mental model is shaped, resulting in input that does not match the robot's way of processing~\citep{kriz2010fictional}. Henceforth, we refer to a \emph{non-functional mental model} in cases where the internal representation about the robot's functionality differs from reality in a way that leads the user to generate incorrect input and to be unable to comprehend the behavior of the robot. If the user is aware of the functionality and limitations of a robot, we refer to a \emph{functional mental model}. We argue that a functional user's mental model of the robot reduces erroneous human-robot interactions.

One approach for the user to gain a functional mental model is to convey the robot's internal architecture. To achieve this goal, two main factors need to be considered. First, the architecture of the robot itself needs to be designed in a way, that even non-expert users can understand its functionality. Second, knowledge about the architecture needs to be conveyed to the user comprehensively. To achieve the best result, one has to balance between functionality and comprehensibility. The development of a complex control architecture to solve a wide variety of problems is at the cost of comprehensibility for users.

This work investigates ways to increase users' knowledge about the architecture of the robot and its influence on the success of human-robot interactions. To increase the comprehensibility of the robot's inner working, we employ a behavior-based architecture we call \emph{FAMILIAR -- FunctionAl user Mental model by Increased LegIbility ARchitecture}. We chose this architecture because it focuses on legibility for the user, while still abstracting parts. Based on this architecture, we implemented two complementary approaches to increase users' knowledge about the architecture. To evaluate these approaches, we conducted an interactive online user study in a simulation to test whether they increase knowledge about the robot's architecture and whether more architecture knowledge does indeed improve human-robot interactions.

\section{Related Work}
\subsection{Mental Model Improvement}
Communication in human-robot interaction fulfills a crucial role in improving the human's understanding of the robot. In the field of didactics for computer science, researcher differentiate between the \emph{relevance} and the \emph{architecture} of computational artifacts \citep{rahwan2019machine,schulte2018framework}. This differentiation describes the dual nature of such artifacts. While an internal mechanism processes the input to generate a behavior (\emph{Architecture}), this behavior becomes observable from the outside (\emph{Relevance}). Knowledge about the relevance of a robot is the basis for a successful interaction. However, in erroneous human-robot interactions, knowledge about the architecture is needed to comprehend the source of the problem.

When interacting with a robot, it is of considerable importance for the user to comprehend the course of the interaction \citep{Wortham2017} and whether an error occurred while executing a task~\citep{BERT, kwon2018expressing}. In many instances, particular actions are only understandable if the overarching goal is known. Therefore, a robot should not only communicate which action it executes but also what goal should be achieved by this action~\citep{huang2019enabling, kaptein2017personalised}.

The execution of actions is preceded by a series of perceptions by the robot which led to the decision. For the user to comprehend, whether a wrong action was executed based on design errors or because the perception process failed, the decision process should also be communicated~\citep{Breazeal, Thomaz09, Otero08}.

While various studies were conducted to measure the influence of communication strategies on users' attitudes towards robots, the mental model was rarely investigated explicitly. Previous work showed na\"ive users can comprehend the architectural concepts of a robot. Furthermore, the comprehension depends on the familiarity and observability of concepts \citep{lukas2020robots}.

\subsection{Robot Control Architecture}
For a robot to interact with the world, sensors need to observe the environment. Based on the perception of the robot, the control module defines which actuators will interact with the world to produce a certain behavior. As this robot control module forms the basis for the robot's interaction behavior, the user's perception and comprehension of it have a tremendous influence on the success of the interaction. As previous work showed, the widely used concept of the \emph{state machine} for robot control is incomprehensible for na\"ive users~\citep{lukas2020robots}. State machines are composed of states and connections between them. A transition from one state to another depends on the state outcome for a given input. Furthermore, in robotic applications, state machines often allow for concurrency~\citep{bohren2010smach}. This highly interlaced structure makes it difficult for na\"ive users to trace the decision-making process, which is especially crucial in case of errors.

Various architectures for robot control use complex control structures to achieve goal-oriented behavior. While these allow for solving challenging tasks, the comprehension by na\"ive users suffers. This is primarily due to the frequently used, hierarchical structure of the planning process~\citep{bryson2001intelligence, colledanchise2017behavior, erol1996hierarchical, peterson1977petri}.

In contrast, behavior-based controls are more reactive in their traditional form~\citep{michaud2016behavior}. They define a set of modules (called behaviors), which are composed of expected sensory input as the trigger and behavioral patterns that achieve a certain goal. These behavioral patterns, in turn, produce an output on actuators~\citep{maes1990learning}. Even though the reactivity reduces the applicability for complex tasks, it increases the legibility. This is achieved by the tight coupling of sensors and their impact on executed behaviors. In contrast, state machines mostly run through an extensive process of state transitions to determine the execution of an action.

\subsection{Knowledge Comprehension}
To be useful for na\"ive users, in the long run, robots need to be flexible and adjustable. To achieve this, non-expert users need to be able to adjust and program the behavior of robots. Out of this need, the field of End-User Development (EUD) emerged \citep{paterno2017new}. To enable novice users to program the behavior of robots, many tools use a visual interface \citep{CORONADO2020100970}. These interfaces allow the user to define the behavior of robots via drag-and-drop of behavioral primitives \citep{huang2017code3, 5326209}. One important difference between the various visual programming software modules is the abstraction of the low-level architecture of the robot. A trade-off between user experience, or simplicity of the software, and the abstraction and deception of the robot architecture must be made \citep{DAGIT2006302}. In our work, we want to impart the architecture of the robot as accurately as possible, while still being understandable. Therefore, our interface is designed towards transparency, taking into account the loss of user satisfaction.

\section{Hypotheses}
\label{sec:hypotheses}
With the goal to improve human-robot interactions through increased knowledge about the architecture of the robot, we hypothesize the following:

\subsection{Hypothesis 1}
A visualization that imparts the inner processes of the robot will increase users' knowledge about the processes of the robot control system.

\subsection{Hypothesis 2}
Through visually programming the behavior of the robot, the users' structural knowledge about the robot's control system will increase.

\subsection{Hypothesis 3}
Insights about the architecture of the robot control system will reduce user induced errors in interactions with a robot.

\section{Methodology}
To design an appropriate robot control architecture, a trade-off between functionality and comprehensibility needs to be made. \citet{diprose2017designing} suggests multiple abstraction levels for robot architectures. This work separates social interaction into five abstraction levels. Namely, these are hardware, algorithm, social and emergent primitives and methods of controlling these primitives. Hardware primitives include the low level hardware, e.g. laser scanner. The algorithm primitives describe algorithms, such as speech recognition. The next higher level of abstraction are the social primitives. These primitives are reusable units for social interaction (e.g. speaking). Building on this are the emergent primitives, which are functions that emerge through the combination of social primitives. The highest abstraction, the methods of controlling primitives, describes how the lower level abstractions are controlled. The conclusion of their investigations suggests, that the abstraction level of social primitives is most suitable for programming robot social interactions. While our research focuses to convey the architecture of the robot to the user, we followed this decision. 

In the following, we will first describe the architecture, which consists of a behavior-based method for controlling primitives (\cf~\cref{sec:architecture}). For our first approach to improve the understandability of the architecture, we developed a dynamic visualization that communicates how the social primitives are used by the architecture (\cf~\cref{subsec:UserInterface}). The second approach, the visual programming of the robot, is described in \cref{subsec:tutorial}.

\subsection{FAMILIAR -- Architecture}
\label{sec:architecture}
The human-robot interaction failure taxonomy by~\citep{honig2018understanding} describes various sources of errors. According to their taxonomy, faulty interactions can either be based on technical issues of the robot or incorrect input by the user. In both cases, the problem typically affects the behavior of the robot and, by this, becomes observable. Therefore, users need to understand how the behaviors of the robot are triggered. Based on this, our control system was designed with regard to two essential factors. The system should be capable to act in a dynamic, fast-changing environment, as well as being comprehensible for na\"{i}ve users. Because the comprehensibility of the architecture was more important, we reduced the complexity of the system for better legibility. We named the architecture \emph{FAMILIAR -- FunctionAl user Mental model by Increased LegIbility ARchitecture}. The system also incorporates a general visualization to make the architecture observable for users. This visualization forms the basis of our dynamic visualization. We developed our behavior system in python 2.7 \citep{van1995python} using the Robot Operating System (ROS) \citep{ros} as middleware.

\begin{figure}
    \centering
    \includegraphics[width=.5\textwidth]{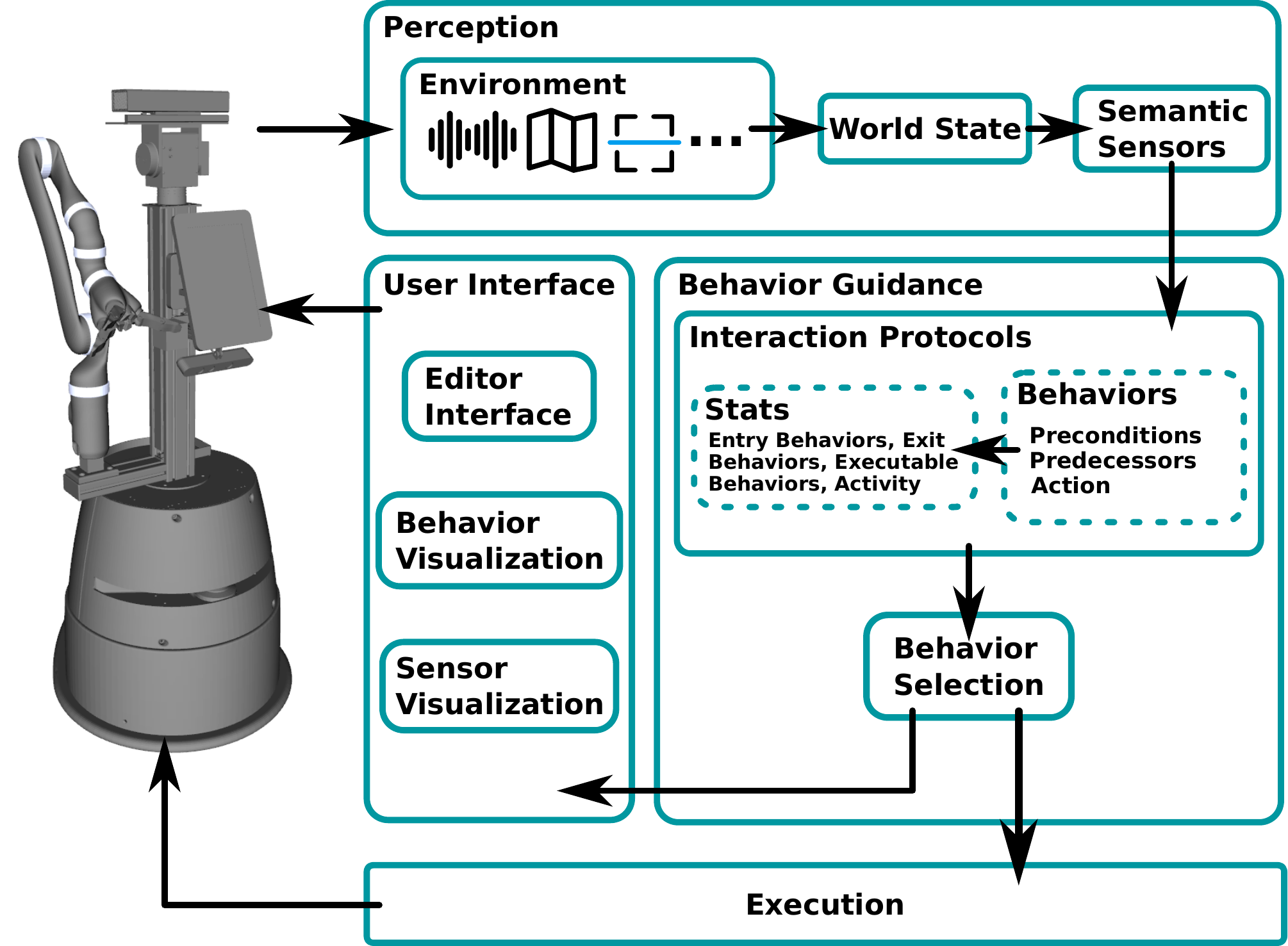}
    \caption{Overview of the robot control architecture.}
    \label{fig:architecture}
\end{figure}

\subsubsection{General Structure}
The general structure of our behavior system is displayed in~\cref{fig:architecture}. The architecture controls the behaviors of the robot by two primary components: the \texttt{Perception} and the \texttt{Behavior Guidance}. The \texttt{Perception} perceives the environment with sensors and extracts higher-level \emph{semantic sensors}. The \texttt{Behavior Guidance} contains the selection process of behaviors to execute. Behaviors to execute are determined by the current set of semantic sensor values. The selected behavior for execution is sent to the \texttt{Execution} component, which executes low-level actions to generate the desired behavior.

\subsubsection{Perception}
\label{subsec:PerAndMem}
To perceive the world, robotic systems need to process through various stages. First, the low-level hardware primitives need to observe the world and represent them in numbers. These observations are then used by algorithm primitives to interpret the observations. While all these operations are important, they are difficult for users to understand. More importantly, if an error occurs in one of these steps, non-expert users would not be able to counteract the problem. Therefore, a more abstract and comprehensible representation of the robot's perception is needed. \citet{saunders2015teach} used so-called \emph{semantic sensors} in their architecture. This type of sensor abstracts the low-level perception and extracts semantically simpler observations of the environment. In that way, users can more easily understand what the robot observed.

Values of semantic sensors are extracted in three steps. First, the robot uses an arbitrary number of hardware sensors together with software components to perceive the world. This can be, for example, a camera for object recognition or a microphone for speech recognition. The perceptions are sent to a memory module, where they are combined with a set of sensors that represent the world (henceforth referred to as "world state"). This world state is used to extract and update semantic sensors. These semantic sensors are then transmitted to the \emph{Behavior Guidance} module. For example, a sensor module that recognizes people would return an array of bounding boxes, which denotes the locations of people as seen by the camera. A semantic sensor that extracts a user comprehensible meaning might be a sensor that denotes a Boolean value if a person is visible or not.

\subsubsection{Behavior Guidance}
\label{subsec:BehaviorGuidance}
The \texttt{Behavior Guidance} process is at the core of our architecture. We use the word \emph{guidance} as behaviors are not only executed based on the current set of semantic sensors. The interaction context (represented as \texttt{Interaction Protocols}) further guides the selection process. The overall guidance component was developed as simple as possible, to be comprehensible for na\"{i}ve users. When errors in the interaction occur, the user should be able to trace the error back to its origin. With more complex robot control designs, the process of error tracing becomes too complex for non-expert users. For example, \citet{saunders2015teach} integrated a hierarchical task network planner (SHOP2) \citep{nau1999shop} to plan subsequent actions to achieve more complex behaviors of the robot.

Our system specifies an \texttt{Interaction Protocol (IP)} as a data structure to group behaviors that need to be executed to fulfill the goal of the interaction. The \texttt{IP} structure allows for more goal-specific interactions which, in turn, supports comprehension for users. An \texttt{IP} is a hierarchical structure and
consists of one behavior that starts the protocol, an arbitrary number of behaviors that can relate to each other, and one or more exit behaviors to end the \texttt{IP}. The data structure also stores information about the execution status of behaviors, as well as a priority value that determines the rank compared to other protocols. \texttt{IPs} with a higher priority value are preferred. Moreover, it is possible to switch from an active \texttt{IP} to a higher prioritized one and later switch back. That way, timing-sensitive goals can be achieved faster.

A \emph{behavior} is described by \emph{preconditions} and \emph{predecessors} that need to be satisfied for execution, and the \emph{action} (with its parameters) that is triggered by the behavior. Preconditions specify desired values of semantic sensors for the behavior to be executable. In addition to preconditions, where semantic sensors are compared to desired values, a behavior can also define a set of predecessors that need to be completed before the behavior can be executed.

When a precondition of a behavior is satisfied and all predecessors are completed, a behavior can be executed. The \emph{Behavior Selection} process determines the behavior to execute (\cf~\cref{subsec:BehaviorSelection}). Executing it triggers the corresponding action. An action is defined by the process triggered in the \texttt{Execution} component and its parameters. The corresponding parameters can be either static values (specified in a configuration file) or dynamic values that depend on the current world state. For example, a static parameter value for a navigation action could be the goal \emph{kitchen}. In contrast, a dynamic parameter could define the goal as the current position of the interaction partner.
The triggered action is transmitted to the \texttt{Execution} component. To achieve the goal of the behavior action, the \texttt{Execution} component executes multiple low-level actions. After the execution of the action is completed, the corresponding behavior is marked as finished.

\subsubsection{Behavior Selection}
\label{subsec:BehaviorSelection}
After semantic sensors are updated, multiple behaviors from different \texttt{IPs} might be executable. Our system does not allow concurrent execution of behaviors. That way, users merely need to focus on one behavior, which improves comprehensibility and failure detection. Therefore, a selection process has to determine the subsequent behavior to execute. This task of guiding the behavior execution is carried out by a higher-level process. This process determines which \texttt{IP} is active (and therefore preferred), and which behavior within the active \texttt{IP} should be executed.

The selection process starts with an update of the current world state. Based on the world state, the affected semantic sensors are updated. Preconditions of behaviors are subscribed to their corresponding semantic sensors. Therefore, an update of semantic sensors also leads to an update of preconditions. If a precondition is fulfilled by the update, the status of the corresponding behavior changes to \emph{executable}.

After all semantic sensors are updated and all executable behaviors are determined, the selection process decides what behavior should be executed. In the first step, the process selects an \texttt{IP} from which a behavior should be executed. In general, the \texttt{IP} that is currently active and includes executable behaviors is preferred. The sole exception to this is the priority of \texttt{IPs}. A higher prioritized \texttt{IP} is preferred to an active \texttt{IP}. This is important to prefer time-sensitive tasks. When an \texttt{IP} is selected, the behavior within the protocol needs to be determined. If the protocol is inactive yet, only the entry behavior can be executed. Otherwise, multiple behaviors may be executable. In that case, behaviors that have predecessors that were executed last are prioritized. Otherwise, behaviors are selected based on their position within the \texttt{IP}. E.g., behaviors that were defined first are also selected first.

\subsection{Architecture Visualization}
\label{subsec:UserInterface}
Along with the behavior architecture, we also developed a graphical user interface to convey each part of the architecture. The goal was to convey the inner working of the robot as accurately as possible, while still being abstract enough to be understandable. To investigate our second hypothesis (\cf~\cref{sec:hypotheses}), the user interface was developed in two versions. The first version only displayed the static information of the architecture, i.e. the components and structures as described above. Based on this basic visualization, we developed an enhanced version, which also displays the dynamic process information.

\begin{figure}
    \centering
    \begin{minipage}[t]{0.4\textwidth}
    \centering
    \includegraphics[width=0.9\linewidth]{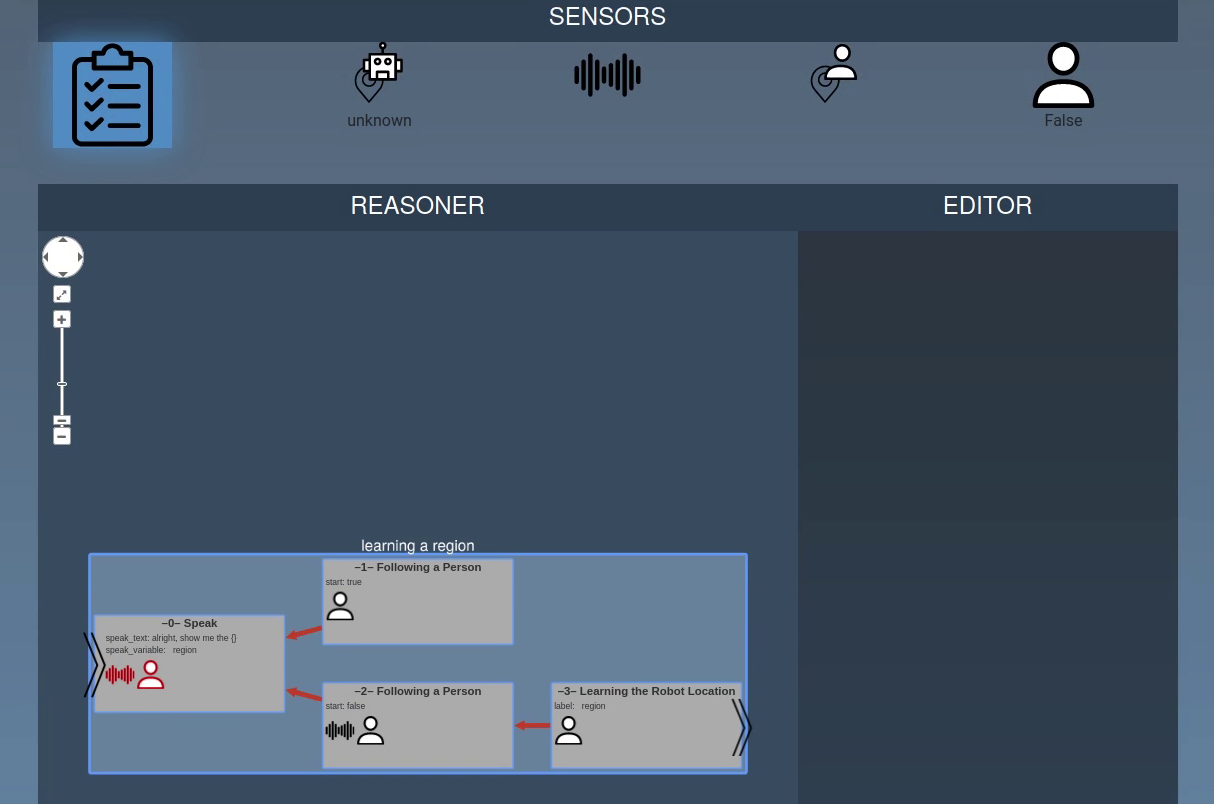}
    \caption{Architecture Visualization Overview. The sensors are displayed at the top. In the bottom left, the interaction protocols and behaviors are displayed. To the right, details of behaviors are displayed. In addition, new behaviors are defined there.}
    \label{fig:reasoner_overview}
    \end{minipage}\qquad
    \begin{minipage}[t]{0.4\textwidth}
    \centering
    \includegraphics[width=1\linewidth]{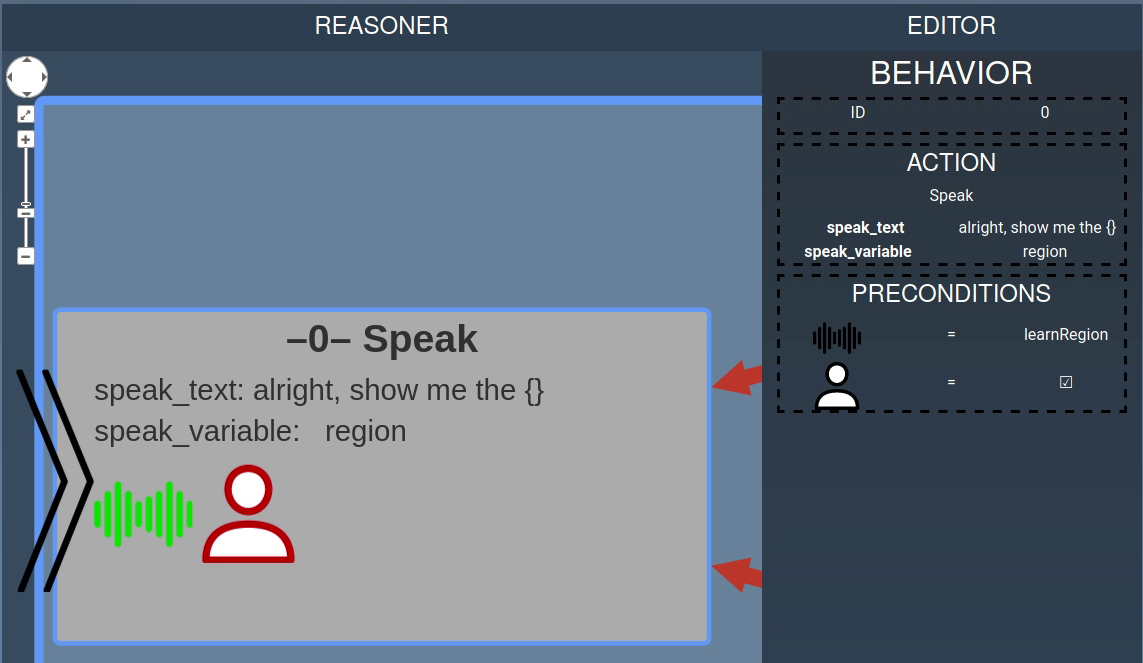}
    \caption{A close-up view of a behavior. When clicking on a behavior, this behavior is zoomed in and details are displayed on the right side.}
    \label{fig:reasoner_behavior}
    \end{minipage}
\end{figure}

Our interface was developed as a web application in JavaScript \citep{flanagan2006javascript} and is divided into two parts (\cf~\cref{fig:reasoner_overview}). These are a visualization of the semantic sensors and their current values, and a visualization of the \texttt{Behavior Guidance} component. All visualization parts were integrated in the web interface of the simulation (\cf~\cref{subsec:web_interface}). In the following, we will discuss each part in more detail.

\subsubsection{Sensor Visualization}
For a successful human-robot interaction, the user must be able to comprehend and track the perception of the robot. Our behavior system extracts semantic sensor values that abstract from the low-level perception of the robot (\cf~\cref{subsec:PerAndMem}). Semantic sensors are visualized at the top of our user interface (\cf~\cref{fig:reasoner_overview}), and displayed side by side. The basic visualization only displays representative icons for each sensor. The enhanced version also displays the current value of the sensor. 

We used icons to represent each sensor to facilitate a fast perception and understanding. For icons to represent the underlying semantic sensor concisely, the selection has to be made with considerable care. Our strategy was to select similar icons to known interfaces. For example, the recognized speech command could be represented by sound waves. This icon is potentially known from speech recognition systems, such as \emph{Google assist}\footnote{\href{https://assistant.google.com}{https://assistant.google.com} [accessed: 2021-07-05]}. To further improve the understanding of the meaning of each icon, we added descriptive tooltip boxes. The tooltip of an icon was printed above when the mouse hovered over the sensor icon.

\subsubsection{Behavior Visualization}
Our \emph{Behavior Guidance} visualization was displayed at the center of our visualization. We focused on displaying the important parts of the system as simple as possible without simplifying or abstracting the inner working too much. To represent the \texttt{IP} structure with their behaviors, we used a network representation, using the cytoscape.js library~\citep{franz2016cytoscape}. Each \texttt{IP} is drawn as a rectangular node, including its behaviors as children. \texttt{IPs} are vertically aligned, and the names are drawn above the node. The containing behaviors of an \texttt{IP} are aligned horizontally within the node. To display the behaviors in a legible way, we calculated an even distribution of the graphical positions of the behavior nodes within the \texttt{IP} node with the Reingold-Tilford algorithm~\citep{reingold1981tidier}.

The behaviors are likewise drawn like rectangular nodes (\cf~\cref{fig:reasoner_behavior}). Each node includes information about the preconditions that have to be fulfilled, as well as the corresponding action that is triggered by the behavior. To distinguish immediately between the various behaviors, a description of the action that is triggered by the behavior is displayed as the title. The preconditions are visualized with the corresponding semantic sensor icons. To receive more detailed information about a behavior, clicking on the behavior node displays additional information on the right side. In addition, entry behaviors have big double right arrows on the left side, indicating the entry point to the corresponding \texttt{IP}. In contrast, exit behaviors have the arrows on the right side, indicating the exit point of the \texttt{IP}. To express the predecessors of a behavior, arrows are drawn from the behavior node towards the predecessor node.

The enhanced visualization adds coloring and highlighting as mechanisms to indicate the processes of the architecture. For the preconditions, we used colored icons to indicate the satisfaction status. Behaviors of an inactive \texttt{IP} have black colored precondition icons, indicating that the corresponding preconditions are not updated. Otherwise, a fulfilled condition is displayed in green. In contrast, unfulfilled conditions are displayed in red. Predecessor arrows are colored red while the predecessor is unexecuted, otherwise it is colored green.

To highlight the selection process, we used \emph{border} and \emph{background} colors, varying sizes, and zooming in on nodes. Nodes of inactive \texttt{IPs} and behaviors have blue border colors and light gray backgrounds. Active \texttt{IPs} and executable behaviors have green border colors. When a behavior is executed, the background color of the corresponding node turns green and is zoomed in. After the execution of a behavior is finished, the background and border color turn dark gray to indicate this behavior has already been executed in the current interaction. Only after the respective \texttt{IP} finished, the behavior border-color changes back to blue, and the background color changes to light gray again.

\subsection{Visual Programming}
\label{subsec:shapeInteraction}
Our second approach, to improve the users' knowledge about the architecture of the robot, is the active definition of the interaction protocol. Through the process of visual programming, the users need to digest their own role in the interaction, as well as think through the robot's role. Thus, develop a deeper understanding of the robot's architecture.

\begin{figure}
    \centering
    \includegraphics[width=1\textwidth]{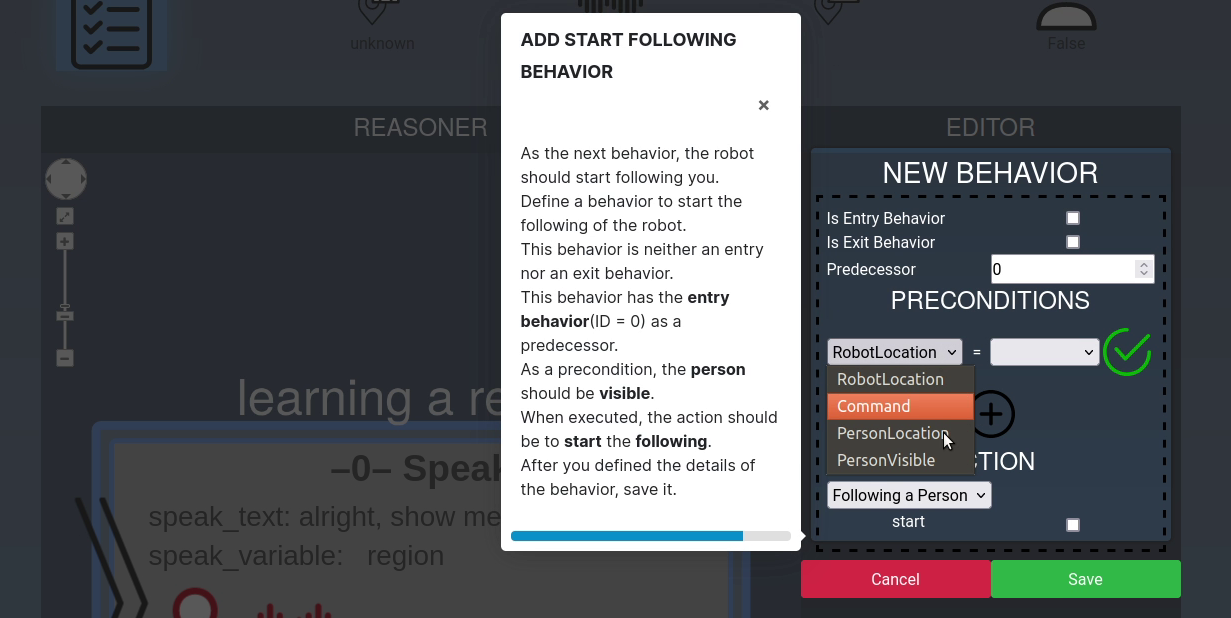}
    \caption{The behavior editor interface with a tutorial step. The tutorial step describes how the behavior should be defined. In the interface, all parts of a behavior can be defined.}
    \label{fig:editor_interface}
\end{figure}

For this, we developed an editor interface to add new \texttt{IP}s and define the containing behaviors with their preconditions and actions. This editor interface is structured similarly to fill out a form. The most important part of this is the behavior definition (\cf~\cref{fig:editor_interface}). To define if this behavior is an \emph{entry}-- or \emph{exit} behavior, it can be selected by checkboxes. The predecessor can be selected by entering the ID of the preceding behavior. The preconditions are defined by selecting the corresponding sensors and their expected values in drop-down menus. Similarly, the action with its parameters is added. 

To guide participants of the study through the process of defining an \texttt{IP}, we developed an interactive tutorial. In the tutorial, participants had to read text boxes with information and had to interact with the editor interface. In this way, we were able to introduce the interface, as well as ensure that all participants define the same \texttt{IP} for the following interaction. Each step of the tutorial was only shown once. The information, presented in the tutorial, was only descriptive of what is shown and which value an input field needs. Therefore, no additional architectural knowledge about the robot was communicated.

The tutorial steps to define the first behavior were designed highly detailed, while the other behaviors were only one tutorial step each, describing what the behavior should look like. In that way, we wanted the participants to not only follow each step, but elaborate on the final interaction protocol.

\subsection{Simulation}
To evaluate our developed architecture and investigate the hypothesis (\cf~\cref{sec:hypotheses}) we designed an online simulation to conduct a human-robot interaction user study. The simulation was developed with Gazebo \citep{koenig2004design} and ROS. For users to interact with the simulation remotely, we implemented a web visualization using gzweb\footnote{\href{http://gazebosim.org/gzweb}{http://gazebosim.org/gzweb} [accessed: 2021-07-05]}. For easy deployment, the simulation was built as a docker \citep{merkel2014docker} image and ran on university servers.

\subsubsection{Server infrastructure and communication with simulation}
Our infrastructure used one main server connected to the internet and multiple backend servers to run an instance of the simulation each. The main server was configured as a reverse proxy to distribute incoming connections to the simulation server. To ensure that each simulation instance was only used by one client, we limited incoming connections to each backend server to one. In addition, we configured the reverse proxy to connect each client always to the same backend server. In addition to the reverse proxy, the main server also ran a web server to provide the resources of the study website.

\subsubsection{Scenario}
\label{subsec:scenario}
\begin{figure}
    \centering
    \includegraphics[width=.5\textwidth]{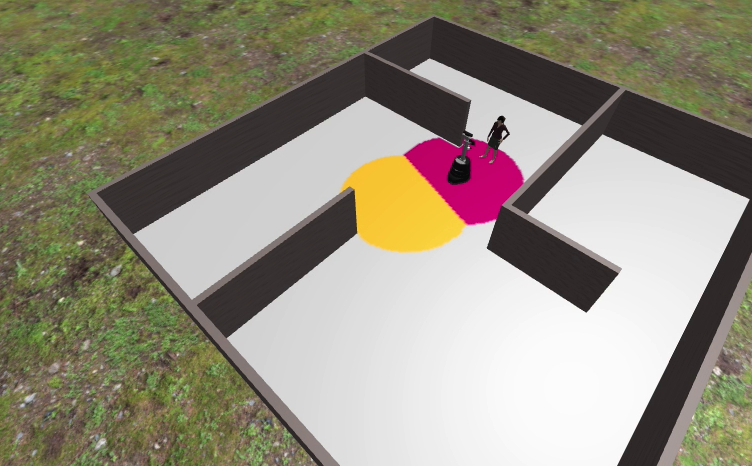}
    \caption{Example of an apartment with two regions taught to the robot. The classification of the robot is indicated on the floor. The yellow region is the kitchen, the purple region is the entrance and the white areas are unclassified.}
    \label{fig:region_learning}
\end{figure}

The simulation and its web interface were developed for a region learning scenario. The goal of the human-robot interaction was to teach the robot about three regions in an apartment. The region learning was realized with a scikit\footnote{\href{https://scikit-learn.org/stable/}{https://scikit-learn.org} [accessed: 2021-07-05]} implementation of the k-nearest neighbor algorithm. Therefore, for learning a region the x-y~coordinate had to be stored together with the region name as the label. The taught regions and the resulting segmentation of the apartment by the robot were visualized with a color texture on the floor of the apartment. An example of how a classified apartment could have looked like is shown in \cref{fig:region_learning}.

To trigger the robot to learn a new region, the user had to move the avatar, representing an interacting user, in sight of the robot. Afterward, a specific command, together with the label of the region, had to be provided. While in a real interaction this command would be given verbally and, thus, have to be recognized by an ASR component, in our online version we used written text that the participant had to specify via a keyboard.
This command triggered the robot to follow the avatar. The user then had to move the avatar around the apartment to guide the robot to the region to learn. When the avatar and the robot arrived at the location to learn, the user had to give the command that they arrived. This triggered the robot to learn the current x-y~location of the robot, together with the label as class. Afterward, the user had to repeat the interaction for the other regions.

For this scenario, we developed several sensors and actuators for the robot. To allow the user to interact with the robot, we equipped the robot with the ability to recognize the avatar of the user and to interpret natural language. To simulate how person recognition would be in the real world, the robot was only able to recognize the avatar around a certain radius from itself. For this, we calculated the Euclidean distance between the current position of the robot and the avatar. This distance determined, whether the avatar is in sight of the robot. This represented approximately the real behavior of the person recognition component, which requires users to be well visible and close enough. For the natural language understanding, we used the \emph{Snips NLU} \citep{coucke2018snips} project. We trained the NLU module to understand more commands than those necessary to achieve the interaction goal. The module also supported the recognition of slight variations of the intended commands.
This allowed us to differentiate, later, if participants of the study only applied the exclusion procedure or understood how to interact with the robot. To interact with the environment, the robot was able to verbally answer the user, via a chat box. Moreover, the robot was able to navigate through the apartment and follow the avatar.

\subsubsection{Web-Interface}
\label{subsec:web_interface}
\begin{figure}
    \centering
    \includegraphics[width=1\textwidth]{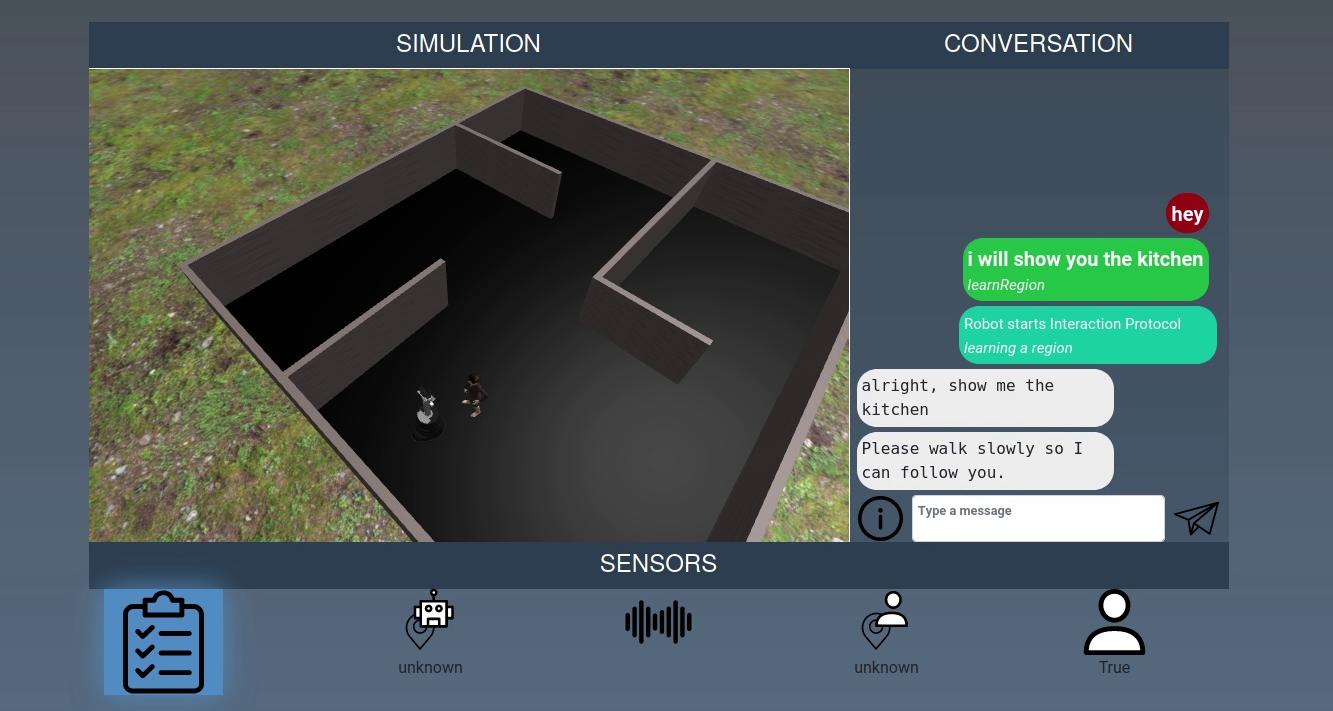}
    \caption{The Simulation Web-Interface. The simulation visualization is displayed in the top left. On the right side, the chat-box interface to communicate with the robot is displayed. Below, the \emph{FAMILIAR}-architecture visualization is located.}
    \label{fig:web_interface}
\end{figure}

The Web-Interface to interact with the simulation consisted of five parts (\cf~\cref{fig:web_interface}). In the top left corner was the graphical interface of the gazebo simulation located. On the right side was the Chat-Box interface. The visual components of the control architecture were displayed below the simulation and Chat-Box interface.

\paragraph{Gazebo-Simulation}
The visualization of the gazebo simulation was realized with gzweb. The camera view was fixed at one location and was oriented towards the apartment. Therefore, the user was able to see the avatar, the robot, and the whole apartment. The height of the walls of the apartment together with the height of the camera was adjusted to prevent the users' view from being blocked. To reduce the amount of resources that need to be loaded, and to reduce the calculation time of the simulation, we excluded any furnishings.

The user was able to move the avatar, by clicking in the simulation window. The 2D position of the click event was then projected on the floor plane of the simulation. The avatar's position was updated with the new position. Therefore, the avatar did not move continuously towards the new location, but directly appear there. Because this mechanism sometimes resulted in the avatar falling over, the physics calculations of the avatar were disabled.

To prevent the avatar to appear outside the apartment, in a wall, or too close to the robot, the position of the click-event was checked to be free of obstacles and within the borders of the apartment. In case the user clicked in an unreachable position, a pop-up window appeared, which informed the user that this location is unreachable. 
(The starting positions of the robot and the avatar were so far apart, that the avatar was not in sight of the robot.)

\paragraph{Chat-Box}
To verbally interact with the robot, we developed a chat-box window. Because we aimed at easy accessibility, the design was inspired by common messengers, such as WhatsApp\footnote{\href{https://www.whatsapp.com/?lang=en}{www.whatsapp.com} [accessed: 2021-07-05]} or Facebook Messenger\footnote{\href{https://www.messenger.com/}{www.messenger.com} [accessed: 2021-07-05]}. The main window displayed the communication history, while the bottom window consisted of elements to write new messages.

The messages of the user were displayed in speech bubbles on the right side. Because the robot could only understand certain commands, the color and content of these speech bubbles depended on the content of the message. If a message could not be interpreted by the robot, the speech bubble was colored \emph{red} and the message was displayed in it. In the case, that the message could be interpreted by the robot, the speech bubble was colored \emph{green}. In Addition, the content of the message, as well as the classified intent by the robot, was displayed in the speech bubble. The answers of the robot were displayed in a white speech bubble on the left side. In addition to the verbal communication of the user and the robot, the chat-box window also displayed the (de-)activation events of the robot control architecture.

The bottom window, to write new messages, consisted of three elements. At the center of this window, the message could be entered in an input field. This message could be sent by clicking on the \emph{send}-button on the right side. In addition, an \emph{information}-button was displayed on the left side. Clicking this button displayed a pop-up window, presenting the intents the robot was able to understand, as well as example sentences for each intent.

\subsection{User Study}
The conducted user study was carried out online using the online platform prolific\footnote{\href{www.prolific.co}{www.prolific.co} [accessed: 2021-07-05]}. Based on our two complementary approaches, we employed a 2-by-2 between-subjects study design, with the two dimensions being participating in a visual programming tutorial before the interaction (VP, no VP) and seeing a dynamic visualization during the interaction (DV, no DV). 
Therefore, participants in the \texttt{Baseline} condition were neither shown the process visualization, nor did participants complete the tutorial. Participants in the \texttt{DV} (Dynamic Visualization) condition were only shown the process visualization, while participants in the \texttt{VP} (Visual Programming) condition only completed the tutorial. Participants in the \texttt{VP+DV} condition completed the tutorial and were shown the process visualization.

\begin{figure}
    \centering
    \includegraphics[width=1\textwidth]{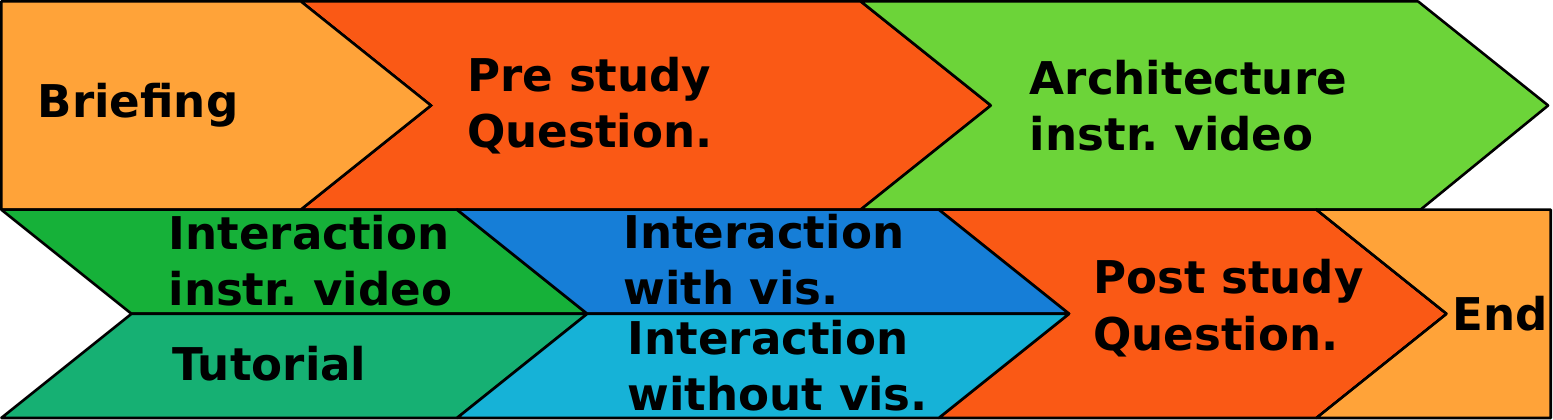}
    \caption{Study schedule. First, participants were briefed and had to answer preceding questionnaires. Afterwards, the architecture instruction video was shown. Depending on the condition, participants completed the tutorial or watched an interaction instruction video. In the main part, participants interacted with the robot to achieve a joint goal. At the end, various questionnaires were collected.}
    \label{fig:study_schedule}
\end{figure}

Acquired participants were forwarded to our study website, where they were assigned to one of the four conditions. The study procedure can be seen in \cref{fig:study_schedule}. First, all participants were shortly briefed about the structure of the study and the estimated time scope. Afterward, their gender, age, and prolific-id were gathered.

To measure the technical affinity of participants, we administered the ATI \citep{ati} questionnaire. Moreover, we measured the memorability with a word memory test \citep{green1996word}. In this test, each participant was shown thirteen word items for ten seconds. They were asked to memorize as many word items as they could. Afterwards, participants were asked to recall and list the memorized word items.

Participants were then shown an instruction video, introducing the simulation and the robot control architecture (\cf~\cref{subsec:intro_video}). Depending on the condition, participants were forwarded to one of two stages. The \texttt{Baseline} and \texttt{DV} conditions were shown another video, explaining at the \emph{relevance} level, how to achieve the interaction goal. In contrast, the \texttt{VP} and \texttt{VP+DV} conditions followed the tutorial to program the behaviors.

In the next phase of the study, all participants interacted with the robot, trying to accomplish the interaction goal (\cf~\cref{subsec:scenario}). To have the same amount of time for each participant, we limited the interaction with the robot to thirty minutes. After the interaction, participants were forwarded to the knowledge questionnaires. Participants who did not achieve the interaction goal within the time limit were asked why they could not achieve the goal before they were also forwarded to the knowledge questionnaires.

To receive more subjective ratings about the system, after the knowledge questionnaires, we collected parts of the Godspeed questionnaire \citep{GOD} and the System-Usability-Scale (SUS) \citep{SUS}. We limited the Godspeed questionnaire to the scales of \emph{Anthropomorphism}, \emph{Likability} and \emph{Perceived Intelligence}. These scales were most relevant to rate the technical system. To end the study, participants were thanked for their participation and forwarded to prolific for their payment.

\subsubsection{Introduction Video}
\label{subsec:intro_video}
Based on our hypotheses (\cf~\cref{sec:hypotheses}), the goal of the introduction video was to inform participants about the architecture of the robot as well as the simulation. The total length of the video was 8:25 minutes. The video was designed as a series of questions and answers together with corresponding video material. The questions were asked by a novice. Thus, questions were asked by someone who is not aware of how the simulation or the robot works. The answers were given by an expert who introduces all important aspects. In that way, we explained all important parts naturally in a conversation instead of an enumeration of facts. The questions and answers were spoken as well as transcribed and displayed. All important aspects of the architecture were introduced together with their visualization in the web interface. The simulation interface is shown in the center of the video. At the bottom of the video, icons for the novice and the expert, as well as the transcription of the spoken text, are displayed.

\subsubsection{Interaction Briefing}
\label{subsec:tutorial}
While the \texttt{VP} and \texttt{VP+DV} conditions went through the tutorial steps, the other conditions were shown an interaction video. This video showed the resulting interaction protocol and described which steps were to follow to teach the robot. This way, participants of all conditions were exposed to the way in which they could interact with the robot, before interacting with it. Thus, the tutorial only influenced how intensively participants interacted with the low-level parts of the architecture.

\subsubsection{Measures}
\label{subsec:know_quest}

\paragraph{Architectural Knowledge}
To measure the resulting architectural knowledge of each participant, we asked several multiple-choice questions about the robot's functionality. Each question consisted of three statements, where only one statement was correct. The question categories for the \emph{structural} knowledge were \texttt{Sensor}, \texttt{IP}, \texttt{Behavior}, \texttt{Precondition}, \texttt{Action}, and \texttt{Predecessor}. The \emph{process} knowledge was asked in the \texttt{Process} category. The knowledge needed to answer all questions was provided in the introduction video (\cf~\cref{subsec:intro_video}). Therefore, all conditions were given the same foundation to answer these questions. Our approaches of dynamic visualization and visual programming only helped consolidate the knowledge. The results of this measurement was used to investigate if the dynamic visualization (H.1) and the visual programming (H.2) improved the architectural knowledge. Additionally, it was used to analyze if an increase in users' knowledge improves the interaction (H.3).

\paragraph{Interaction Success}
To further analyze the interaction of the participants, we calculated several key figures from the log data of the simulation interaction. To get an overall impression of how successful participants were in each condition, we divided participants into two groups based on whether they achieved the interaction goal: interactions of participants who taught all three regions to the robot within the time limit, were classified as \texttt{successful}, otherwise the interactions were \texttt{unsuccessful}. This grouping was used together with the architectural knowledge to answer hypothesis 3.

\paragraph{Interaction Failures}
We also evaluated mistakes of the user during the interaction. For this, we counted the number of wrong commands provided by the user. We classified a command as wrong, if the given command was unneeded to reach the interaction goal. Therefore, a command could be wrong even though the robot was able to interpret it. Moreover, we counted the amount of times users moved the avatar out of sight of the robot. Thus, indicating how aware participants were about the perception limitations of the robot. These measurements were used to further investigate the influence of knowledge on the interaction success.

\paragraph{Godspeed, SUS, and ATI}
As already mentioned above, we also collected data of the Godspeed questionnaire (Anthropomorphism, Likability, Perceived Intelligence) and the System-Usability-Scale. In contrast to the objective measurement above, the results from these questionnaires were used to get insights of the subjective rating of the interaction.

\section{Results}
We conducted our study with 85 participants, of which 4 were excluded due to technical problems. The remaining 81 participants were randomly assigned to the four conditions (\texttt{Baseline}: n=20, \texttt{DV}: n=21, \texttt{VP}: n=20, \texttt{DV+VP}: n=20). The average age of participants was 35 (SD=9.53) years. Overall, 30 female, 50 male and 1 diverse persons participated. A Kruskal-Wallis test \citep{Kruskal} showed no significant difference between the conditions regarding age (H(4)=3.9091, p=0.2714), gender (H(4)=0.9606, p=0.8108), ATI (H(4)=0.8627, p=0.8344) or word recall (H(4)=0.5732, p=0.9025).

For a first overview, we analyzed the architectural knowledge between the conditions. The data of each knowledge category was checked towards normality via a Shapiro-Wilk~\citep{SHAPIRO1965} test. In the case of normally distributed data, we applied a one-way ANOVA. Otherwise, we used a Kruskal-Wallis test. When the one-way ANOVA indicated a significance, we conducted a Tukey-HSD~\citep{tukeyHSD} posthoc test. In the case of the Kruskal-Wallis test, we used a follow-up Posthoc Dunn~\citep{dunn} test. The results can be seen in \cref{tab:knowledge_cond} and \cref{fig:quest}. The results showed, that the overall knowledge only differed significantly between the \texttt{Baseline}(M=39.29, SD=10.16) and \texttt{VP+DV}(M=52.44, SD=19.5) (p=0.0582) condition. Additionally, the \texttt{Behavior} knowledge differed between the \texttt{Baseline}(M=35.0, SD=27.84) and the \texttt{VP}(M=70.0, SD=29.15) (p=0.0011) and \texttt{VP+DV}(M=57.5, SD=36.31) (p=0.0349) conditions, and between the \texttt{DV}(M=45.24, SD=30.49) and \texttt{VP} (p=0.0190) conditions. 

\begin{table}[]
    \centering
    \caption{Statistical tests for architectural knowledge of conditions. }
    \begin{tabular}{ll}
        \toprule
        \begin{tabular}[t]{l S[table-format=3.4] S[table-format=1.4]}
             \multicolumn{3}{c}{\textbf{Sensor}}\\
             \rowcolor{gray!15}
             Test & {statistic} & {p-value}\\
             \rowcolor{white}
             Shapiro-Wilk & 0.8132 & 0.0000\\
             Kruskal-Wallis & 1.4472 & 0.6945\\
             \midrule
             \multicolumn{3}{c}{\textbf{IP}}\\
             \rowcolor{gray!15}
             Test & {statistic} & {p-value}\\
             \rowcolor{white}
             Shapiro-Wilk & 0.6355 & 0.0000\\
             Kruskal-Wallis & 0.7772 & 0.8549\\
             \midrule
             \multicolumn{3}{c}{\textbf{Behavior}}\\
             \rowcolor{gray!15}
             Test & {statistic} & {p-value}\\
             \rowcolor{white}
             Shapiro-Wilk & 0.8002 & 0.0000\\
             Kruskal-Wallis & 12.0788 & \textbf{0.0071}\\
             \multicolumn{3}{l}{Posthoc Dunn}\\
             Condition 1    & {Condition 2} & {p-value}\\
             \midrule
             \rowcolor{gray!15}
                                            & \texttt{DV}                 & 0.3362\\
                                            & \texttt{VP}                 & \textbf{0.0011}\\
             \multirow{-3}{*}{\texttt{Baseline}} & \texttt{VP+DV}            & \textbf{0.0349}\\
             \rowcolor{white}
                                                 & \texttt{VP}                 & \textbf{0.0190}\\
             \multirow{-2}{*}{\texttt{DV}}      & \texttt{VP+DV}            & 0.2406\\
             \rowcolor{gray!15}
             \texttt{VP}               & \texttt{VP+DV}                 & 0.2472\\
             \rowcolor{white}
             \midrule
             \multicolumn{3}{c}{\textbf{Precondition}}\\
             \rowcolor{gray!15}
             Test & {statistic} & {p-value}\\
             \rowcolor{white}
             Shapiro-Wilk & 0.8718 & 0.0000\\
             Kruskal-Wallis & 4.7849 & 0.1882\\
        \end{tabular}
        &
        \begin{tabular}[t]{l S[table-format=3.4] S[table-format=1.4]}
             \multicolumn{3}{c}{\textbf{Action}}\\
             \rowcolor{gray!15}
             Test & {statistic} & {p-value}\\
             \rowcolor{white}
             Shapiro-Wilk & 0.7986 & 0.0000\\
             Kruskal-Wallis & 2.8695 & 0.4122\\
             \midrule
             \multicolumn{3}{c}{\textbf{Predecessor}}\\
             \rowcolor{gray!15}
             Test & {statistic} & {p-value}\\
             \rowcolor{white}
             Shapiro-Wilk & 0.5647 & 0.0000\\
             Kruskal-Wallis & 2.2326 & 0.5256\\
             \midrule
             \multicolumn{3}{c}{\textbf{Process}}\\
             \rowcolor{gray!15}
             Test & {statistic} & {p-value}\\
             \rowcolor{white}
             Shapiro-Wilk & 0.7954 & 0.0000\\
             Kruskal-Wallis & 2.2326 & 0.5256\\
             \midrule
             \multicolumn{3}{c}{\textbf{All}}\\
             \rowcolor{gray!15}
             Test & {statistic} & {p-value}\\
             \rowcolor{white}
             Shapiro-Wilk & 0.9848 & 0.4527\\
             1-Way ANOVA & 2.9281 & \textbf{0.0389}\\
             \multicolumn{3}{l}{Tukey-HSD}\\
             Condition 1    & {Condition 2} & {p-value}\\
             \midrule
             \rowcolor{gray!15}
                                            & \texttt{DV}                 & 0.7712\\
                                            & \texttt{VP}                 & 0.1049\\
             \multirow{-3}{*}{\texttt{Baseline}} & \texttt{VP+DV}            & \textit{0.0582}\\
             \rowcolor{white}
                                              & \texttt{VP}                 & 0.4912\\
             \multirow{-2}{*}{\texttt{DV}}    & \texttt{VP+DV}            & 0.3424\\
             \rowcolor{gray!15}
             \texttt{VP}               & \texttt{VP+DV}                 & 0.9\\
        \end{tabular}\\
        \bottomrule
    \end{tabular}
    \label{tab:knowledge_cond}
\end{table}

\subsection{H1: Process Visualization improves process knowledge}
Our first hypothesis stated, that the visualization about the processes of the architecture will increase the process knowledge about this architecture. To get insights about this hypothesis, we analyzed the knowledge questionnaire scores. We tested between each condition and also the \texttt{DV} and \texttt{DV+VP} conditions pooled together. No significant differences between the conditions could be found. Therefore, our first hypothesis could not be confirmed.

\begin{table}[]
    \caption{Statistical tests for the amount of visualization usages. P-values $\leq$ 0.1 are indicated by \textit{italic} font, P-values $\leq$ 0.05 are indicated in \textbf{bold} font.}
    \centering
    \begin{tabular}{l S[table-format=1.4] S[table-format=1.4]}
    \toprule
    \rowcolor{gray!15}
    Test & {Statistic} & {p-value}\\
    \rowcolor{white}
    Shapiro-Wilk & 0.7989 & 0.0000\\
    Kruskal-Wallis & 8.3282 & \textbf{0.0397}\\
    \midrule
    Condition 1    & {Condition 2} & {p-value}\\
    \rowcolor{gray!15}
                                    & \texttt{DV}                 & \textbf{0.0491}\\
    \rowcolor{gray!15}
                                    & \texttt{VP}                 & 0.7356\\
    \rowcolor{gray!15}
    \multirow{-3}{*}{\texttt{Baseline}} & \texttt{VP+DV}            & 0.6239\\
    \rowcolor{white}
    \multirow{2}{*}{\texttt{DV}}   & \texttt{VP}                 & \textbf{0.0225}\\
                                    & \texttt{VP+DV}            & \textbf{0.0145}\\
    \rowcolor{gray!15}                                
    \texttt{VP}               & \texttt{VP+DV}                 & 0.8809\\
    \end{tabular}
    \label{tab:lookat_vis}
\end{table}

\subsection{H2: Visual programming improves structural knowledge about the robot}
\begin{figure}
    \centering
    \includegraphics[width=0.5\textwidth]{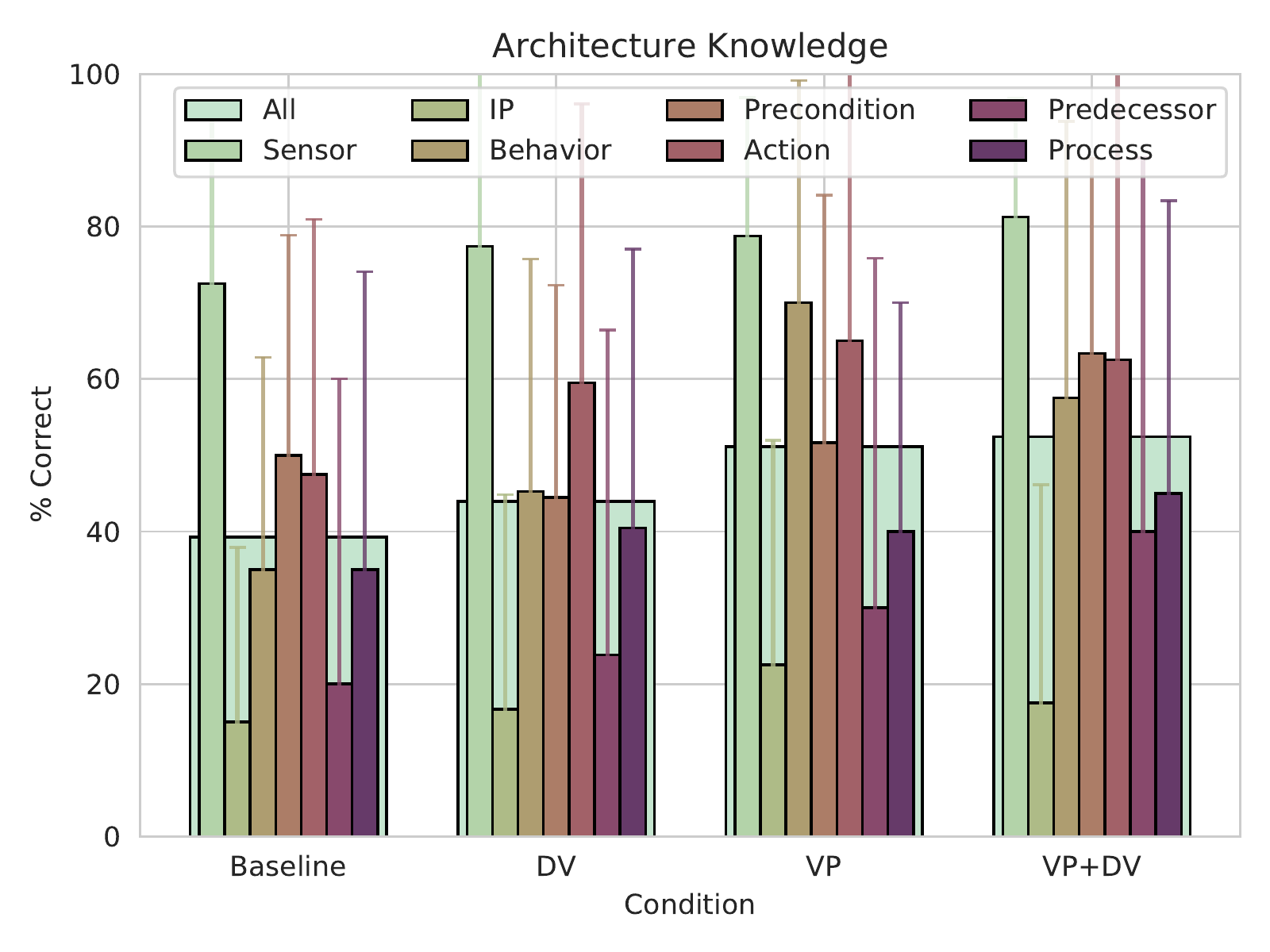}%
    \includegraphics[width=0.5\textwidth]{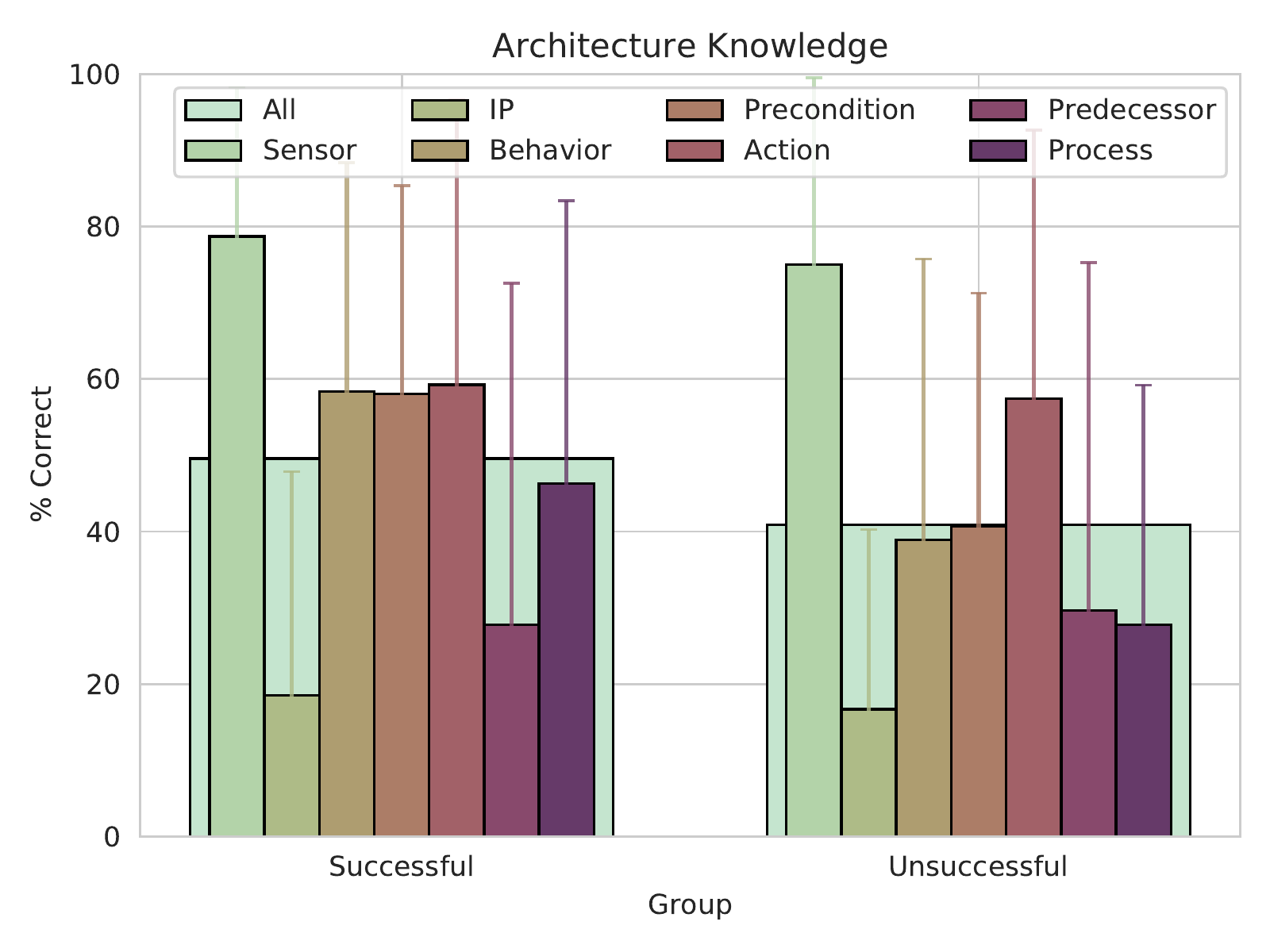}
    \caption{Architecture Knowledge for each condition (left); and groups \texttt{successful} and \texttt{unsuccessful} (right)}
    \label{fig:quest}
\end{figure}

In contrast to our process visualization, we hypothesized that visually programming the interaction would improve the structural knowledge about the robot which corresponds to the knowledge categories \texttt{Sensor}, \texttt{IP}, \texttt{Behavior}, \texttt{Precondition}, \texttt{Action}, and \texttt{Predecessor}. To evaluate this, we compared the architectural knowledge of the \emph{visual programming} (\texttt{VP}, \texttt{VP+DV}) conditions with the conditions \emph{without visual programming}(\texttt{Baseline}, \texttt{DV}). A Shapiro-Wilk normality check with a follow-up T-test showed that participants in the visual programming conditions had significantly more knowledge about the overall architecture (statistic=-2.8264, p=0.0060). Additionally, a Mann-Whitney-U test showed that knowledge about the \texttt{Behavior} concept (statistic=520.50, p=0.0018) was also increased. Moreover, there was a tendency towards more knowledge of the \texttt{Precondition} concept in the VP condition (\cf~\cref{tab:knowledge_quest_cond} and \cref{fig:quest}). Even though the visual programming did not improve knowledge about all concepts, we conclude that our second hypothesis can be partly confirmed.

\begin{table}[]
    \centering
    \caption{Statistical analysis of the architectural knowledge for groupings of no \texttt{VP} vs. \texttt{VP}. Tests were either Mann-Whitney-U (MWU) or T-Test (T) depending on the results from the Shapiro-Wilk normality check. P-Values $\leq$ 0.1 are indicated by \textit{italic} font, P-Values $\leq$ 0.05 are indicated in \textbf{bold} font.} 
    \begin{tabular}{l S[table-format=3.4] c}
    \toprule
    \multirow{2}{*}{Category} & \multicolumn{2}{c}{no \texttt{VP} vs. \texttt{VP}}\\\cline{2-3}
     & {statistic} & {p-value}\\
    \rowcolor{gray!15}
    Sensor \tiny{(MWU)} & 765.0 & 0.5772 \\
    IP \tiny{(MWU)} & 768.0 & 0.5505\\
    \rowcolor{gray!15}
    Behavior \tiny{(MWU)} & 520.50 & \textbf{0.0018}\\
    Prec. \tiny{(MWU)} & 650.0 & \textit{0.0894}\\
    \rowcolor{gray!15}
    Action \tiny{(MWU)} & 693.50 & 0.1996\\
    Pred. \tiny{(MWU)} & 713.0 & 0.1977\\
    \rowcolor{gray!15}
    Process \tiny{(MWU)} & 754.0 & 0.5039\\
    All \tiny{(T)}  & -2.8264 & \textbf{0.0060}\\
    \bottomrule
    \end{tabular}
    \label{tab:knowledge_quest_cond}
\end{table}

\subsection{H3: Architecture knowledge improves human-robot interaction}
\begin{figure}
    \centering
    \includegraphics[width=.5\linewidth]{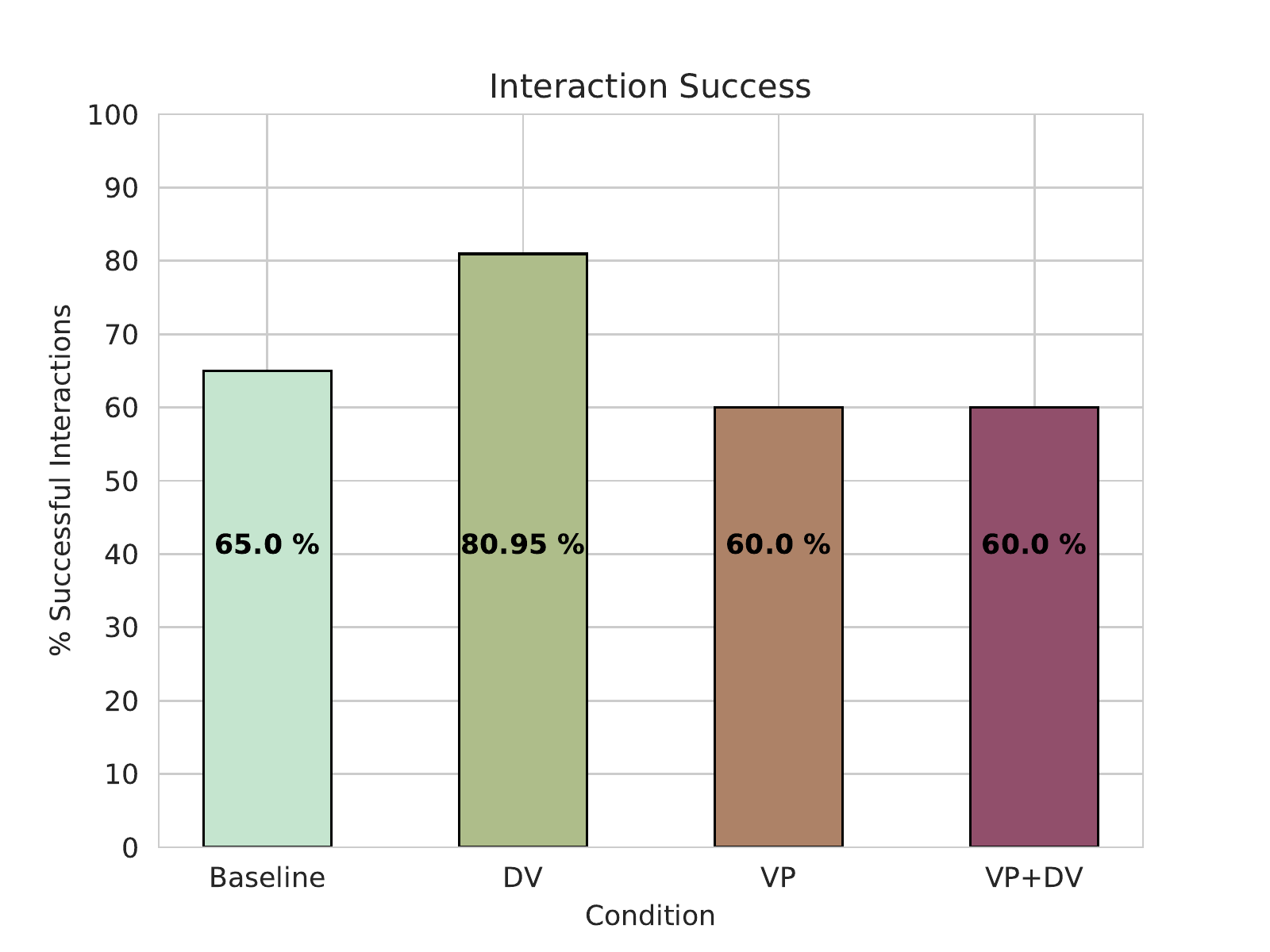}%
    \includegraphics[width=.5\linewidth]{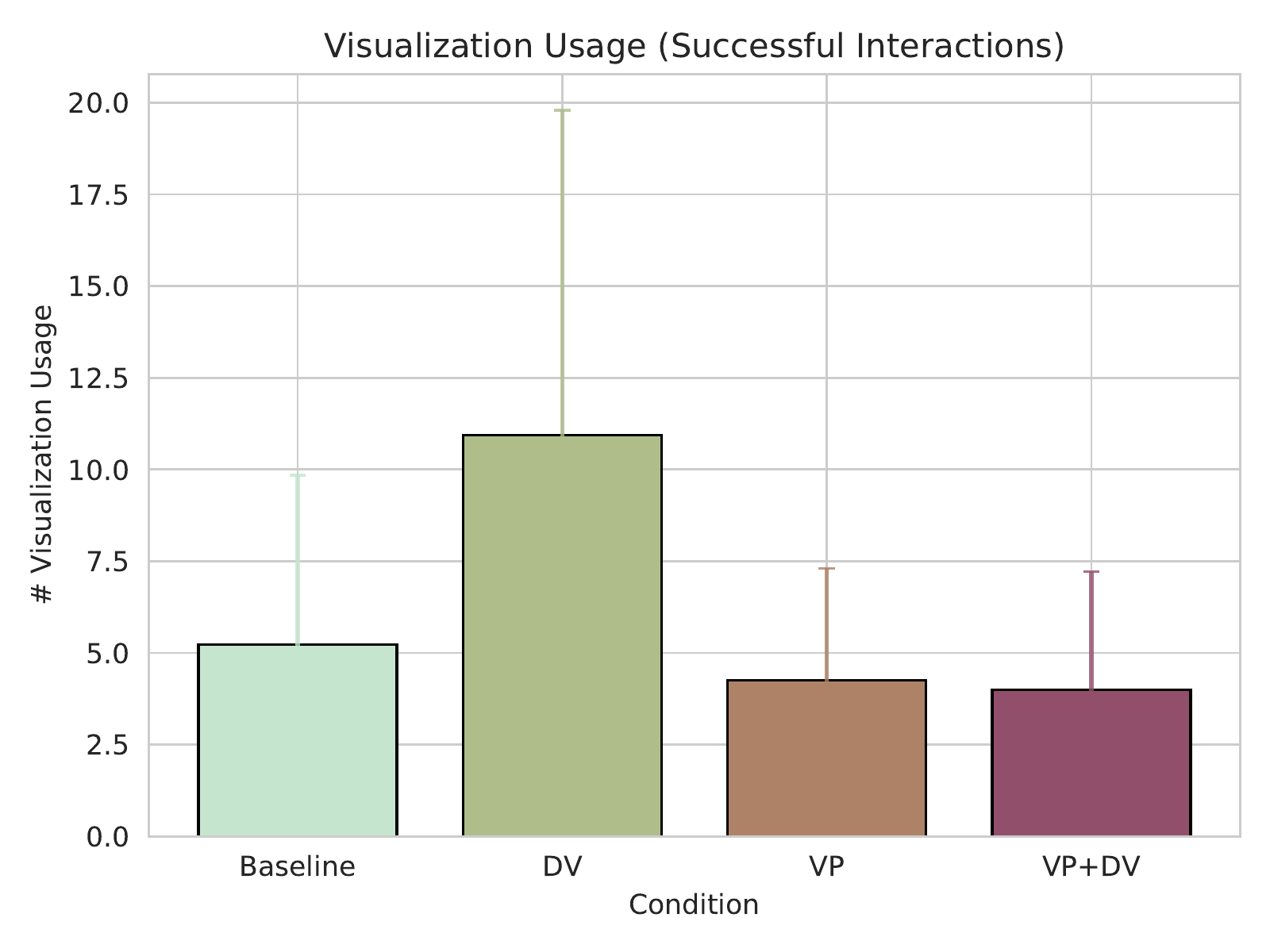}
    \caption{(left) Amount of successful interactions per condition; (right) Amount of visualization usages per condition.}
    \label{fig:vis_num}
\end{figure}

Our third hypothesis stated, that knowledge about the robot control architecture would improve the human-robot interaction. Our first measure concerned the amount of successful interactions per group. As it can be seen in \cref{fig:vis_num}, 80.95\% of the participants in the \texttt{DV} condition achieved the interaction goal, while only 65\% of the \texttt{Baseline} and 60\% of the \texttt{VP} and \texttt{VP+DV} conditions achieved the goal. To further investigate the influence of architectural knowledge on the interaction, we compared the \texttt{successful} and \texttt{unsuccessful} interaction groups. We applied a Shapiro-Wilk normality check and a Mann-Whitney-U test for non-normal distributed data and a T-Test otherwise. The results can be seen in \cref{tab:knowledge_quest_group} and \cref{fig:quest}. Overall, the results show that participants who were able to achieve the interaction goal had more knowledge about the architecture of the robot. A closer look at each category reveals that knowledge about \texttt{Behavior}, \texttt{Precondition} and \texttt{Process} were higher.

Another key measure was the number of wrong commands given. To compare participants who used a high number of wrong commands with those who used fewer wrong commands, we calculated a median split for the number of wrong commands. Again, we compared the answer score of each category and the overall score with a Mann-Whitney-U or T-Test. Results can be seen in \cref{tab:knowledge_quest_group}. They showed a tendency for the overall knowledge(statistic=1.9199, p=0.0585), the \texttt{IP}(statistic=677.0, p=0.0986) and \texttt{Behavior}(statistic=988.50, p=0.0789) categories. This indicates that participants with a low number of wrong commands had more knowledge about the \texttt{IP} and \texttt{Behavior} categories, and more overall architecture knowledge.
Moreover, the \texttt{Process}(statistic=1068.0, p=0.0116) knowledge was significantly higher for those with a lower number of used wrong commands.

We also investigated how often the participants moved the avatar out of sight of the robot. For this, we compared the visualization conditions with the conditions without a visualization. We only analyzed the successful interactions, because some participants, who did not achieve the interaction goal, never moved the avatar. Thus, they also never moved the avatar actively out of sight of the robot. A Mann-Whitney-U test showed a tendency of decrease of the number of avatar loosing between \emph{no visualization} (M=9.16, SD=4.95) and \emph{visualization} (M=6.72, SD=4.30) conditions (statistic=468.0, p=\textit{0.0676}).

Based on these results, we conclude that our third hypotheses can be confirmed. We further could observe that certain concepts of our architecture were of particular importance for an improved interaction. Additionally, this is the reason, why the visual programming improved the knowledge about the architecture but could not improve the interaction itself.

\begin{table}[]
    \centering
    \caption{Statistical analysis of the architectural knowledge for groupings of \emph{Successful Interaction}, \emph{Wrong Commands}. Tests were either Mann-Whitney-U (MWU) or T-Test (T) depending on the results from the Shapiro-Wilk normality check. P-Values $\leq$ 0.1 are indicated by \textit{italic} font, P-Values $\leq$ 0.05 are indicated in \textbf{bold} font.} 
    \begin{tabular}{l S[table-format=3.4] c S[table-format=3.4] c}
    \toprule
    \multirow{2}{*}{Category} & \multicolumn{2}{c}{Successful Interaction} & \multicolumn{2}{c}{Wrong Commands}\\\cline{2-5}
     & {statistic} & {p-value} & {statistic} & {p-value}\\
    \rowcolor{gray!15}
    Sensor \tiny{(MWU)} & 770.50 & 0.6565  & 762.50 & 0.5599\\
    IP \tiny{(MWU)} & 729.0 & 0.9951 & 677.0 & \textit{0.0986}\\
    \rowcolor{gray!15}
    Behavior \tiny{(MWU)} & 945.0 & \textbf{0.0168} & 988.50 & \textit{0.0789}\\
    Prec. \tiny{(MWU)} & 935.0 & \textbf{0.0289} & 983.50 & 0.1024\\
    \rowcolor{gray!15}
    Action \tiny{(MWU)} & 753.0 & 0.7997 & 925.50 & 0.2581\\
    Pred. \tiny{(MWU)} & 715.50 & 0.8676 & 875.0 & 0.5098\\
    \rowcolor{gray!15}
    Process \tiny{(MWU)} & 925.0 & \textbf{0.0344} & 1068.0 & \textbf{0.0116}\\
    All \tiny{(T)} & 2.2487 & \textbf{0.0273} & 1.9199 & \textit{0.0585}\\
    \bottomrule
    \end{tabular}
    \label{tab:knowledge_quest_group}
\end{table}

\subsection{Subjective Ratings of the Interaction}
\begin{figure}
    \centering
    \includegraphics[width=0.6\textwidth]{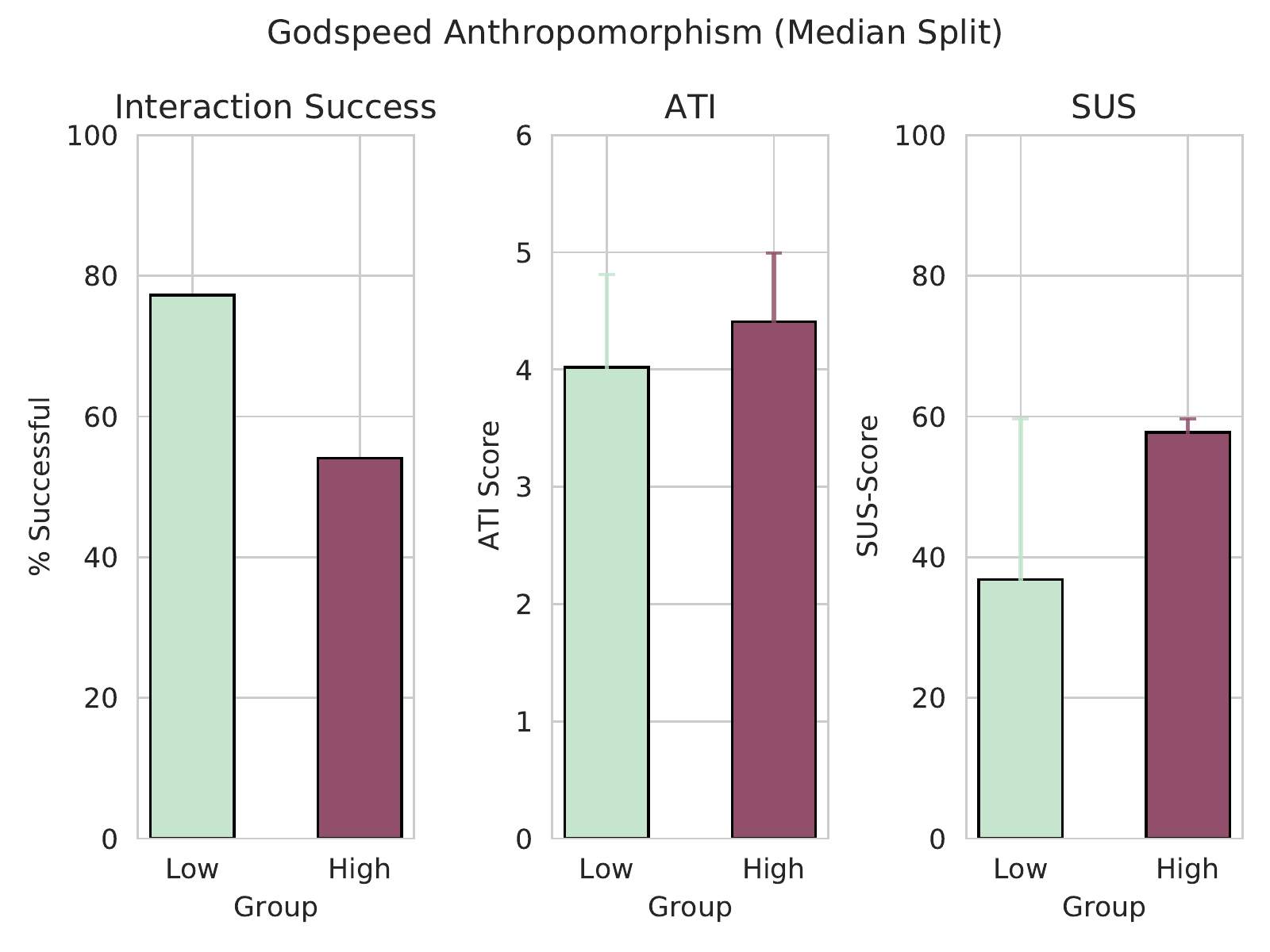}
    \caption{Comparison of godspeed anthropomorphism median split for \emph{Interaction Success}, \emph{ATI} and \emph{SUS}}
    \label{fig:god_anthro}
\end{figure}

In addition to the knowledge questionnaires, we also collected data of the godspeed questionnaire and the SUS. While we could not find any differences for the \emph{likability} or \emph{perceived intelligence}, differences for the \emph{anthropomorphism} score could be found. In general, in social robotics it is seen as a positive result if users anthropomorphize a robot as this indicates that the robot is seen as a social entity. However, it is unknown how this affects the user's reasoning about a non-human entity such as a robot. To investigate which influence the anthropomorphisation of participants had on the success of the interaction, we applied a median split on the anthropomorphism score. We then compared the interaction success of participants with lower anthropomorphism score with the interaction success of participants with a higher anthropomorphism score (\cf~\cref{fig:god_anthro}). A Shapiro-Wilk normality check (statistic=0.9096, p=0.0000) with a Mann-Whitney-U test (statistic=488.50, p=\textbf{0.0518}) revealed that participants with a lower anthropomorphism score were more likely able to achieve the interaction goal (\cf~\cref{tab:anthro}).

We also compared the SUS and ATI for the anthropomorphism median split. A Mann-Whitney-U test revealed that higher anthropomorphism is related to a higher rating of the system usability (statistic=417.50, p=\textbf{0.0002}), as well as a higher ATI score (statistic=602.50, p=\textbf{0.0449}). Additionally, a Mann-Whitney-U test showed a significantly lower SUS for participants in the VP conditions (Statistic= 1089.50, p=\textbf{0.0110}).

\begin{table}[]
    \centering
    \caption{Statistical tests of anthropomorphism median split for \emph{Interaction Success}, \emph{ATI} and \emph{SUS}. P-Values $\leq$ 0.1 are indicated by \textit{italic} font, P-Values $\leq$ 0.05 are indicated in \textbf{bold} font.}
    \begin{tabular}{l c c}
        \toprule
        \multicolumn{3}{c}{\textbf{Interaction Success}}\\
        \rowcolor{gray!15}
        Test & Statistic & P-Value  \\
        Shapiro-Wilk & 0.9096 & \textbf{0.0000}\\
        Mann-Whitney-U & 488.50 & \textbf{0.0158}\\
        \midrule
        \multicolumn{3}{c}{\textbf{ATI}}\\
        \rowcolor{gray!15}
        Test & Statistic & P-Value  \\
        Shapiro-Wilk & 0.9694 & \textbf{0.0000}\\
        Mann-Whitney-U & 602.50 & \textbf{0.0449}\\
        \midrule
        \multicolumn{3}{c}{\textbf{SUS}}\\
        \rowcolor{gray!15}
        Test & Statistic & P-Value  \\
        Shapiro-Wilk & 0.9603 & \textbf{0.0000}\\
        Mann-Whitney-U & 417.50 & \textbf{0.0002}\\
        \bottomrule
    \end{tabular}
    \label{tab:anthro}
\end{table}

\subsection{Qualitative Observations}
In addition to the quantitative results, we also observed some qualitative results. While $\frac{2}{3}rd$ of the participants were able to achieve the interaction goal, the participants who did not achieve the interaction goal revealed some insights on their problems. One observation was that from the 27 participants who did not achieve the interaction goal within the 30-minute time limit, only $\frac{1}{3}rd$ were able to teach at least one region. From the 18 participants who did not even teach one region, 14 participants were in one of the \texttt{VP} conditions. This result is a consequence of the fact, that participants of these conditions either took too long to successfully complete the visual programming sequence or were not even able to complete the visual programming in the time limit. Upon closer investigation of the log-data of the visual programming sequence revealed that participants had problems to specify the \texttt{Predecessor} and add \texttt{Preconditions} to the behavior.

Problems in the interaction with the robot itself were mostly based on providing the correct commands. Even though the VP interface provided a pop-up window to retrieve commands that can be interpreted by the robot, some/many participants did not use this. Instead, they seemed to assume a chatbot like interaction, where the robot always answers. Another common problem was, that participants correctly triggered the "Following" behavior of the robot, to guide it to the region to learn, but then never gave the second command to trigger the learning of this region.

\section{Discussion}
The analysis of our user study revealed some interesting insights into the effects, knowledge has on interaction performance. Participants, who achieved the interaction goal, had significantly more knowledge about the robot's architecture. Upon closer inspection, the knowledge about \texttt{Behavior}, \texttt{Precondition} and \texttt{Process} were critical for the success of the interaction. These concepts, together with the \texttt{Action} concept, build the core of our architecture. While knowledge about the \texttt{Action} concept did not differ between \texttt{successful} and \texttt{unsuccessful} participants, the questions were answered as correct as the \texttt{Behavior} and \texttt{Precondition} questions by the \emph{successful} participants. Additionally, we not only observed that architectural knowledge improves the interaction, but is a factor that makes the interaction possible in the first place. From the participants who could not achieve the interaction goal within the 30-minute time limit, $\frac{2}{3}$ not even taught one region. The investigation of the log data revealed, that these participants had general knowledge gaps. Not only did they provide the wrong commands, but communicated with the robot as it was a chatbot. Something participants, who achieved the interaction goal, did not. This indicates problems of the mental model. We assume that the wrong implications of the visual interface are the reason for this misconception. Because chat-boxes are often used to communicate with other human beings or a chatbot, wrong assumptions about the functionality were created. Thus, a different interface might have worked better.

While the concept of visual programming is not new, the resulting understanding of the system by novice users was merely investigated \citep{CORONADO2020100970}. In this study, we used visual programming as an approach to improve knowledge about the structure of the architecture. The results show that participants of the \texttt{VP} conditions had increased knowledge about the architecture. In addition to the overall knowledge, especially the \texttt{Behavior} knowledge was significantly improved. While our visual programming interface showed positive effects on the knowledge, not all aspects of our architecture could be conveyed. Because additional knowledge about \texttt{Precondition} and \texttt{Process} concepts were critical for the interaction success, we could not observe an increase in the interaction performance for the VP conditions. Additionally, a negative side effect, in the form of a lower SUS rating, could be observed. Therefore, further work needs to investigate how a visual programming interface should be designed to successfully communicate all aspects of the architecture. For this, an increased focus should be put on the \texttt{Precondition} and \texttt{Process} knowledge. We expect that the abstract mechanism of the visual programming, together with the highly flexible way people act, has a major influence on the difficulty to understand these concepts. Even though \texttt{Preconditions} in our architecture are more flexible than in e.g. state machines, they still expect the environment to be in a specific state. In contrast, humans can reason about their environment and other humans on a higher level. To overcome these false assumptions, the visual programming interface should be closely related to the interaction itself. In that way, users might better understand the impact of the concepts on the interaction. 

Although the dynamic visualization of our architecture did not lead to a measurable improvement of the knowledge, participants were yet more likely to achieve the interaction goal. The improved ability to reach the interaction goal indicates improvement of knowledge. Presumably, our knowledge questionnaire did not retrieve this information. The effect of an improved interaction could not be observed for the combination of the VP and the DV in the \texttt{DV+VP} condition. One reason might be a cognitive overload caused by the VP. Presumably, a more accessible visual programming interface together with the dynamic visualization could have improved the interaction.

Previous work showed that visualizing the behavior decision improves the understanding of what the robot does \citep{Wortham2017}. Additionally, \citet{lukas2020robots} showed that the communication of the robots' functionality improves the ability to recognize erroneous interactions in videos. In this work, we have gone a step further, by mediating in-depth knowledge about the architecture of the robot and testing this knowledge in actual human-robot interactions. In this way, we were able to get insights into the specific aspects of a robot control architecture that influence the interaction. Moreover, we enabled persons not only to detect errors in interactions, but also to overcome and reduce these errors and achieve a joint goal with the robot. While the work in \citep{lukas2020robots} is not entirely comparable to ours, only around $\frac{1}{3}$ of participants were able to detect errors based on the way the state machine works. In contrast, we enabled $\frac{2}{3}$ of our participants to achieve an interaction goal.

Besides the influence of the communicated knowledge for improved interaction, the anthropomorphization of the robot influenced the interaction. We observed that participants with less anthropomorphism were more likely to achieve the interaction goal. Similar results were also observed in \citep{lukas2020robots}. In this work, participants with a lower anthropomorphism were more likely to detect errors in a human-robot interaction. 
In general, the question after cause and effect arises. One explanation could be that participants interacted unsuccessfully with the robot because they anthropomorphized the robot. The other direction would be that participants anthropomorphized the robot because the interaction was unsuccessful. We believe that increased anthropomorphization of the robot negatively affected the interaction success. A reason for this might be the difference in the flexible way humans can act, in contrast to the comparatively rigid way of the behavior architecture of the robot. When humans interact with other people, the \emph{theory of mind} area in the brain is activated~\citep{frith2005theory}. The same effect can be observed when interacting with an anthropomorphized robot~\citep{krach2008can}. Using this area of the brain might lead to the incomprehension of how to interact with the robot. This, in turn, results in faulty inputs by the user. Therefore, in contrast to many current approaches, inducing an anthropomorphic image on users \citep{breazeal2016social} can reduce the interaction success.

\section{Conclusion and Future Work}
In this work, we investigated the influence of knowledge about the robot control architecture for human-robot interactions. For this, we employed a behavior-based architecture, which aims at better understandability and transparency. Our study showed that knowledge about the architecture improves human-robot interactions. Moreover, this knowledge only makes it possible to successfully interact with the robot. Therefore, we argue that researcher and engineers should include factors of transparency and comprehensibility in their design processes. These factors can critically influence the success of interactions between users and robots. We also showed that the usage of visual programming can be used to improve the knowledge about architectural aspects of the robot. Moreover, design decisions regarding the anthropomorphisation of robots can negatively influence the interaction.

While the visual programming approach improved the overall knowledge of the architecture, some concepts could not be improved significantly. Therefore, the interaction itself did not improve. One reason might be, that the tutorial was too abstract to understand some aspects of it. Therefore, we will further investigate this approach. One idea could be, to incorporate approaches closer to the interaction. In that way, abstract aspects of the architecture could be directly observed in the interaction.

\section*{Acknowledgements}
We acknowledge the support of the Honda Research Institute Europe GmbH, Carl-Legien-Strasse 30, 63073 Offenbach, Germany. Christiane B. Wiebel-Herboth is employed by the Honda Research Institute Europe GmbH. The remaining authors declare that the research was conducted in the absence of any commercial or financial relationships that could be construed as a potential conflict of interest.
We also acknowledge the financial support of the German Research Foundation (DFG) and the Open Access Publication Fund of Bielefeld University for the article processing charge.

\bibliographystyle{unsrtnat}
\bibliography{references}

\begin{thebibliography}{54}
\providecommand{\natexlab}[1]{#1}
\providecommand{\url}[1]{\texttt{#1}}
\expandafter\ifx\csname urlstyle\endcsname\relax
  \providecommand{\doi}[1]{doi: #1}\else
  \providecommand{\doi}{doi: \begingroup \urlstyle{rm}\Url}\fi

\bibitem[Sendhoff and Wersing(2020)]{sendhoff2020cooperative}
Bernhard Sendhoff and Heiko Wersing.
\newblock Cooperative intelligence-a humane perspective.
\newblock In \emph{2020 IEEE International Conference on Human-Machine Systems
  (ICHMS)}, pages 1--6. IEEE, 2020.

\bibitem[Johnson et~al.(2009)Johnson, Saboe, Prewett, Coovert, and
  Elliott]{johnson2009autonomy}
Ryan~C Johnson, Kristin~N Saboe, Matthew~S Prewett, Michael~D Coovert, and
  Linda~R Elliott.
\newblock Autonomy and automation reliability in human-robot interaction: A
  qualitative review.
\newblock In \emph{Proceedings of the Human Factors and Ergonomics Society
  Annual Meeting}, volume~53, pages 1398--1402. SAGE Publications Sage CA: Los
  Angeles, CA, 2009.

\bibitem[Tsarouhas and Fourlas(2016)]{tsarouhas2016mission}
Panagiotis~H Tsarouhas and George~K Fourlas.
\newblock Mission reliability estimation of mobile robot system.
\newblock \emph{International Journal of System Assurance Engineering and
  Management}, 7\penalty0 (2):\penalty0 220--228, 2016.

\bibitem[Brooks(2017)]{brooks2017human}
Daniel~J Brooks.
\newblock \emph{A human-centric approach to autonomous robot failures}.
\newblock PhD thesis, University of Massachusetts Lowell, 2017.

\bibitem[Steinbauer(2012)]{steinbauer2012survey}
Gerald Steinbauer.
\newblock A survey about faults of robots used in robocup.
\newblock In \emph{Robot Soccer World Cup}, pages 344--355. Springer, 2012.

\bibitem[Honig and Oron-Gilad(2018)]{honig2018understanding}
Shanee Honig and Tal Oron-Gilad.
\newblock Understanding and resolving failures in human-robot interaction:
  Literature review and model development.
\newblock \emph{Frontiers in psychology}, 9:\penalty0 861, 2018.

\bibitem[Staggers and Norcio(1993)]{staggers1993mental}
Nancy Staggers and Anthony~F. Norcio.
\newblock Mental models: concepts for human-computer interaction research.
\newblock \emph{International Journal of Man-machine studies}, 38\penalty0
  (4):\penalty0 587--605, 1993.

\bibitem[Nass et~al.(1994)Nass, Steuer, and Tauber]{nass1994computers}
Clifford Nass, Jonathan Steuer, and Ellen~R Tauber.
\newblock Computers are social actors.
\newblock In \emph{Proceedings of the SIGCHI conference on Human factors in
  computing systems}, pages 72--78, 1994.

\bibitem[Kriz et~al.(2010)Kriz, Ferro, Damera, and Porter]{kriz2010fictional}
Sarah Kriz, Toni~D Ferro, Pallavi Damera, and John~R Porter.
\newblock Fictional robots as a data source in hri research: Exploring the link
  between science fiction and interactional expectations.
\newblock In \emph{19th International Symposium in Robot and Human Interactive
  Communication}, pages 458--463. IEEE, 2010.

\bibitem[Rahwan et~al.(2019)Rahwan, Cebrian, Obradovich, Bongard, Bonnefon,
  Breazeal, Crandall, Christakis, Couzin, Jackson, et~al.]{rahwan2019machine}
Iyad Rahwan, Manuel Cebrian, Nick Obradovich, Josh Bongard, Jean-Fran{\c{c}}ois
  Bonnefon, Cynthia Breazeal, Jacob~W Crandall, Nicholas~A Christakis, Iain~D
  Couzin, Matthew~O Jackson, et~al.
\newblock Machine behaviour.
\newblock \emph{Nature}, 568\penalty0 (7753):\penalty0 477--486, 2019.

\bibitem[Schulte and Budde(2018)]{schulte2018framework}
Carsten Schulte and Lea Budde.
\newblock A framework for computing education: Hybrid interaction system: The
  need for a bigger picture in computing education.
\newblock In \emph{Proceedings of the 18th Koli Calling International
  Conference on Computing Education Research}, pages 1--10, 2018.

\bibitem[Wortham et~al.(2017)Wortham, Theodorou, and Bryson]{Wortham2017}
Robert~H. Wortham, Andreas Theodorou, and Joanna~J. Bryson.
\newblock {Improving robot transparency: Real-Time visualisation of robot AI
  substantially improves understanding in naive observers}.
\newblock \emph{RO-MAN 2017 - 26th IEEE International Symposium on Robot and
  Human Interactive Communication}, 2017-January:\penalty0 1424--1431, 2017.
\newblock \doi{10.1109/ROMAN.2017.8172491}.

\bibitem[{Hamacher} et~al.(2016){Hamacher}, {Bianchi-Berthouze}, {Pipe}, and
  {Eder}]{BERT}
A.~{Hamacher}, N.~{Bianchi-Berthouze}, A.~G. {Pipe}, and K.~{Eder}.
\newblock Believing in bert: Using expressive communication to enhance trust
  and counteract operational error in physical human-robot interaction.
\newblock In \emph{2016 25th IEEE International Symposium on Robot and Human
  Interactive Communication (RO-MAN)}, pages 493--500, 08 2016.
\newblock \doi{10.1109/ROMAN.2016.7745163}.

\bibitem[Kwon et~al.(2018)Kwon, Huang, and Dragan]{kwon2018expressing}
Minae Kwon, Sandy~H Huang, and Anca~D Dragan.
\newblock Expressing robot incapability.
\newblock In \emph{Proceedings of the 2018 ACM/IEEE International Conference on
  Human-Robot Interaction}, pages 87--95, 2018.

\bibitem[Huang et~al.(2019)Huang, Held, Abbeel, and Dragan]{huang2019enabling}
Sandy~H Huang, David Held, Pieter Abbeel, and Anca~D Dragan.
\newblock Enabling robots to communicate their objectives.
\newblock \emph{Autonomous Robots}, 43\penalty0 (2):\penalty0 309--326, 2019.

\bibitem[Kaptein et~al.(2017)Kaptein, Broekens, Hindriks, and
  Neerincx]{kaptein2017personalised}
Frank Kaptein, Joost Broekens, Koen Hindriks, and Mark Neerincx.
\newblock Personalised self-explanation by robots: The role of goals versus
  beliefs in robot-action explanation for children and adults.
\newblock In \emph{2017 26th IEEE International Symposium on Robot and Human
  Interactive Communication (RO-MAN)}, pages 676--682. IEEE, 2017.

\bibitem[Breazeal et~al.(2005)Breazeal, Kidd, Thomaz, Hoffman, and
  Berlin]{Breazeal}
Cynthia Breazeal, Cory Kidd, A.L. Thomaz, Guy Hoffman, and M.~Berlin.
\newblock Effects of nonverbal communication on efficiency and robustness in
  human-robot teamwork.
\newblock pages 708 -- 713, 09 2005.
\newblock ISBN 0-7803-8912-3.
\newblock \doi{10.1109/IROS.2005.1545011}.

\bibitem[{Thomaz} and {Cakmak}(2009)]{Thomaz09}
A.~L. {Thomaz} and M.~{Cakmak}.
\newblock Learning about objects with human teachers.
\newblock In \emph{2009 4th ACM/IEEE International Conference on Human-Robot
  Interaction (HRI)}, pages 15--22, 03 2009.
\newblock \doi{10.1145/1514095.1514101}.

\bibitem[{Otero} et~al.(2008){Otero}, {Alissandrakis}, {Dautenhahn}, {Nehaniv},
  {Syrdal}, and {Koay}]{Otero08}
N.~{Otero}, A.~{Alissandrakis}, K.~{Dautenhahn}, C.~{Nehaniv}, D.~S. {Syrdal},
  and K.~L. {Koay}.
\newblock Human to robot demonstrations of routine home tasks: Exploring the
  role of the robot's feedback.
\newblock In \emph{2008 3rd ACM/IEEE International Conference on Human-Robot
  Interaction (HRI)}, pages 177--184, 03 2008.
\newblock \doi{10.1145/1349822.1349846}.

\bibitem[Hindemith et~al.(2020)Hindemith, Vollmer, G\"opfert, Wiebel-Herboth,
  and Wrede]{lukas2020robots}
Lukas Hindemith, Anna-Lisa Vollmer, Jan~Phillip G\"opfert, Christiane~B.
  Wiebel-Herboth, and Britta Wrede.
\newblock Why robots should be technical: Correcting mental models through
  technical architecture concepts.
\newblock \emph{submitted to Interaction Studies Journal}, 2020.

\bibitem[Bohren and Cousins(2010)]{bohren2010smach}
Jonathan Bohren and Steve Cousins.
\newblock The smach high-level executive [ros news].
\newblock \emph{IEEE Robotics \& Automation Magazine}, 17\penalty0
  (4):\penalty0 18--20, 2010.

\bibitem[Bryson(2001)]{bryson2001intelligence}
Joanna~Joy Bryson.
\newblock \emph{Intelligence by design: principles of modularity and
  coordination for engineering complex adaptive agents}.
\newblock PhD thesis, Massachusetts Institute of Technology, 2001.

\bibitem[Colledanchise and {\"O}gren(2017)]{colledanchise2017behavior}
Michele Colledanchise and Petter {\"O}gren.
\newblock Behavior trees in robotics and ai: an introduction.
\newblock \emph{arXiv preprint arXiv:1709.00084}, 2017.

\bibitem[Erol(1996)]{erol1996hierarchical}
Kutluhan Erol.
\newblock \emph{Hierarchical task network planning: formalization, analysis,
  and implementation}.
\newblock PhD thesis, 1996.

\bibitem[Peterson(1977)]{peterson1977petri}
James~L Peterson.
\newblock Petri nets.
\newblock \emph{ACM Computing Surveys (CSUR)}, 9\penalty0 (3):\penalty0
  223--252, 1977.

\bibitem[Michaud and Nicolescu(2016)]{michaud2016behavior}
Fran{\c{c}}ois Michaud and Monica Nicolescu.
\newblock Behavior-based systems.
\newblock In \emph{Springer handbook of robotics}, pages 307--328. Springer,
  2016.

\bibitem[Maes and Brooks(1990)]{maes1990learning}
Pattie Maes and Rodney~A Brooks.
\newblock Learning to coordinate behaviors.
\newblock In \emph{AAAI}, volume~90, pages 796--802, 1990.

\bibitem[Patern{\`o} and Wulf(2017)]{paterno2017new}
Fabio Patern{\`o} and Volker Wulf.
\newblock New perspectives in end-user development.
\newblock 2017.

\bibitem[Coronado et~al.(2020)Coronado, Mastrogiovanni, Indurkhya, and
  Venture]{CORONADO2020100970}
Enrique Coronado, Fulvio Mastrogiovanni, Bipin Indurkhya, and Gentiane Venture.
\newblock Visual programming environments for end-user development of
  intelligent and social robots, a systematic review.
\newblock \emph{Journal of Computer Languages}, 58:\penalty0 100970, 2020.
\newblock ISSN 2590-1184.
\newblock \doi{https://doi.org/10.1016/j.cola.2020.100970}.
\newblock URL
  \url{https://www.sciencedirect.com/science/article/pii/S2590118420300307}.

\bibitem[Huang and Cakmak(2017)]{huang2017code3}
Justin Huang and Maya Cakmak.
\newblock Code3: A system for end-to-end programming of mobile manipulator
  robots for novices and experts.
\newblock In \emph{2017 12th ACM/IEEE International Conference on Human-Robot
  Interaction (HRI}, pages 453--462. IEEE, 2017.

\bibitem[Pot et~al.(2009)Pot, Monceaux, Gelin, and Maisonnier]{5326209}
E.~Pot, J.~Monceaux, R.~Gelin, and B.~Maisonnier.
\newblock Choregraphe: a graphical tool for humanoid robot programming.
\newblock In \emph{RO-MAN 2009 - The 18th IEEE International Symposium on Robot
  and Human Interactive Communication}, pages 46--51, 2009.
\newblock \doi{10.1109/ROMAN.2009.5326209}.

\bibitem[Dagit et~al.(2006)Dagit, Lawrance, Neumann, Burnett, Metoyer, and
  Adams]{DAGIT2006302}
Jason Dagit, Joseph Lawrance, Christoph Neumann, Margaret Burnett, Ronald
  Metoyer, and Sam Adams.
\newblock Using cognitive dimensions: Advice from the trenches.
\newblock \emph{Journal of Visual Languages \& Computing}, 17\penalty0
  (4):\penalty0 302--327, 2006.
\newblock ISSN 1045-926X.
\newblock \doi{https://doi.org/10.1016/j.jvlc.2006.04.006}.
\newblock URL
  \url{https://www.sciencedirect.com/science/article/pii/S1045926X06000279}.
\newblock Ten Years of Cognitive Dimensions.

\bibitem[Diprose et~al.(2017)Diprose, MacDonald, Hosking, and
  Plimmer]{diprose2017designing}
James Diprose, Bruce MacDonald, John Hosking, and Beryl Plimmer.
\newblock Designing an api at an appropriate abstraction level for programming
  social robot applications.
\newblock \emph{Journal of Visual Languages \& Computing}, 39:\penalty0 22--40,
  2017.

\bibitem[Van~Rossum and Drake~Jr(1995)]{van1995python}
Guido Van~Rossum and Fred~L Drake~Jr.
\newblock \emph{Python reference manual}.
\newblock Centrum voor Wiskunde en Informatica Amsterdam, 1995.

\bibitem[Quigley et~al.(2009)Quigley, Conley, Gerkey, Faust, Foote, Leibs,
  Wheeler, Ng, et~al.]{ros}
Morgan Quigley, Ken Conley, Brian Gerkey, Josh Faust, Tully Foote, Jeremy
  Leibs, Rob Wheeler, Andrew~Y Ng, et~al.
\newblock Ros: an open-source robot operating system.
\newblock In \emph{ICRA workshop on open source software}, volume~3, page~5.
  Kobe, Japan, 2009.

\bibitem[Saunders et~al.(2015)Saunders, Syrdal, Koay, Burke, and
  Dautenhahn]{saunders2015teach}
Joe Saunders, Dag~Sverre Syrdal, Kheng~Lee Koay, Nathan Burke, and Kerstin
  Dautenhahn.
\newblock “teach me--show me”—end-user personalization of a smart home
  and companion robot.
\newblock \emph{IEEE Transactions on Human-Machine Systems}, 46\penalty0
  (1):\penalty0 27--40, 2015.

\bibitem[Nau et~al.(1999)Nau, Cao, Lotem, and Munoz-Avila]{nau1999shop}
Dana Nau, Yue Cao, Amnon Lotem, and Hector Munoz-Avila.
\newblock Shop: Simple hierarchical ordered planner.
\newblock In \emph{Proceedings of the 16th international joint conference on
  Artificial intelligence-Volume 2}, pages 968--973, 1999.

\bibitem[Flanagan(2006)]{flanagan2006javascript}
David Flanagan.
\newblock \emph{JavaScript: the definitive guide}.
\newblock " O'Reilly Media, Inc.", 2006.

\bibitem[Franz et~al.(2016)Franz, Lopes, Huck, Dong, Sumer, and
  Bader]{franz2016cytoscape}
Max Franz, Christian~T Lopes, Gerardo Huck, Yue Dong, Onur Sumer, and Gary~D
  Bader.
\newblock Cytoscape. js: a graph theory library for visualisation and analysis.
\newblock \emph{Bioinformatics}, 32\penalty0 (2):\penalty0 309--311, 2016.

\bibitem[Reingold and Tilford(1981)]{reingold1981tidier}
Edward~M. Reingold and John~S. Tilford.
\newblock Tidier drawings of trees.
\newblock \emph{IEEE Transactions on software Engineering}, \penalty0
  (2):\penalty0 223--228, 1981.

\bibitem[Koenig and Howard(2004)]{koenig2004design}
Nathan Koenig and Andrew Howard.
\newblock Design and use paradigms for gazebo, an open-source multi-robot
  simulator.
\newblock In \emph{2004 IEEE/RSJ International Conference on Intelligent Robots
  and Systems (IROS)(IEEE Cat. No. 04CH37566)}, volume~3, pages 2149--2154.
  IEEE, 2004.

\bibitem[Merkel(2014)]{merkel2014docker}
Dirk Merkel.
\newblock Docker: lightweight linux containers for consistent development and
  deployment.
\newblock \emph{Linux journal}, 2014\penalty0 (239):\penalty0 2, 2014.

\bibitem[Coucke et~al.(2018)Coucke, Saade, Ball, Bluche, Caulier, Leroy,
  Doumouro, Gisselbrecht, Caltagirone, Lavril, et~al.]{coucke2018snips}
Alice Coucke, Alaa Saade, Adrien Ball, Th{\'e}odore Bluche, Alexandre Caulier,
  David Leroy, Cl{\'e}ment Doumouro, Thibault Gisselbrecht, Francesco
  Caltagirone, Thibaut Lavril, et~al.
\newblock Snips voice platform: an embedded spoken language understanding
  system for private-by-design voice interfaces.
\newblock \emph{arXiv preprint arXiv:1805.10190}, pages 12--16, 2018.

\bibitem[Franke et~al.(2019)Franke, Attig, and Wessel]{ati}
Thomas Franke, Christiane Attig, and Daniel Wessel.
\newblock A personal resource for technology interaction: development and
  validation of the affinity for technology interaction (ati) scale.
\newblock \emph{International Journal of Human--Computer Interaction},
  35\penalty0 (6):\penalty0 456--467, 2019.

\bibitem[Green et~al.(1996)Green, Allen, and Astner]{green1996word}
P~Green, LM~Allen, and K~Astner.
\newblock The word memory test: A user’s guide to the oral and
  computer-administered forms, us version 1.1.
\newblock \emph{Durham, NC: CogniSyst}, 1996.

\bibitem[Bartneck et~al.(2009)Bartneck, Kuli{\'c}, Croft, and Zoghbi]{GOD}
Christoph Bartneck, Dana Kuli{\'c}, Elizabeth Croft, and Susana Zoghbi.
\newblock Measurement instruments for the anthropomorphism, animacy,
  likeability, perceived intelligence, and perceived safety of robots.
\newblock \emph{International journal of social robotics}, 1\penalty0
  (1):\penalty0 71--81, 2009.

\bibitem[Bangor et~al.(2008)Bangor, Kortum, and Miller]{SUS}
Aaron Bangor, Philip~T Kortum, and James~T Miller.
\newblock An empirical evaluation of the system usability scale.
\newblock \emph{Intl. Journal of Human--Computer Interaction}, 24\penalty0
  (6):\penalty0 574--594, 2008.

\bibitem[Breslow(1970)]{Kruskal}
Norman Breslow.
\newblock A generalized kruskal-wallis test for comparing k samples subject to
  unequal patterns of censorship.
\newblock \emph{Biometrika}, 57\penalty0 (3):\penalty0 579--594, 1970.

\bibitem[Shapiro and Wilk(1965)]{SHAPIRO1965}
S.~S. Shapiro and M.~B. Wilk.
\newblock An analysis of variance test for normality (complete samples).
\newblock \emph{Biometrika}, 52\penalty0 (3-4):\penalty0 591--611, dec 1965.
\newblock \doi{10.1093/biomet/52.3-4.591}.
\newblock URL \url{https://doi.org/10.1093/biomet/52.3-4.591}.

\bibitem[Abdi and Williams(2010)]{tukeyHSD}
Herve Abdi and Lynne~J Williams.
\newblock Tukey’s honestly significant difference (hsd) test.
\newblock \emph{Encyclopedia of Research Design. Thousand Oaks, CA: Sage},
  pages 1--5, 2010.

\bibitem[Dunn(1964)]{dunn}
Olive~Jean Dunn.
\newblock Multiple comparisons using rank sums.
\newblock \emph{Technometrics}, 6\penalty0 (3):\penalty0 241--252, 1964.

\bibitem[Frith and Frith(2005)]{frith2005theory}
Chris Frith and Uta Frith.
\newblock Theory of mind.
\newblock \emph{Current biology}, 15\penalty0 (17):\penalty0 R644--R645, 2005.

\bibitem[Krach et~al.(2008)Krach, Hegel, Wrede, Sagerer, Binkofski, and
  Kircher]{krach2008can}
S{\"o}ren Krach, Frank Hegel, Britta Wrede, Gerhard Sagerer, Ferdinand
  Binkofski, and Tilo Kircher.
\newblock Can machines think? interaction and perspective taking with robots
  investigated via fmri.
\newblock \emph{PloS one}, 3\penalty0 (7):\penalty0 e2597, 2008.

\bibitem[Breazeal et~al.(2016)Breazeal, Dautenhahn, and
  Kanda]{breazeal2016social}
Cynthia Breazeal, Kerstin Dautenhahn, and Takayuki Kanda.
\newblock Social robotics.
\newblock \emph{Springer handbook of robotics}, pages 1935--1972, 2016.

\end{thebibliography}

\end{document}